\RequirePackage{fix-cm}

\documentclass[smallcondensed]{svjour3} 

\smartqed 

\usepackage{graphicx}
\usepackage{epsfig}
\usepackage{verbatim}
\usepackage{paralist}
\usepackage[utf8]{inputenc}
\usepackage{fixltx2e}
\usepackage{enumitem}
\usepackage{multirow}
\usepackage{hyperref}
\usepackage{listings}
\usepackage{textcomp}
\usepackage{natbib}
\usepackage{url}
\usepackage{fancyvrb}

\journalname{Language Resources and Evaluation}

\AtBeginDocument{
\setlength{\oddsidemargin}{\dimexpr(\paperwidth-\textwidth)/2-1in}
\setlength{\evensidemargin}{\oddsidemargin}
\setlength{\topmargin}{\dimexpr(\paperheight-\textheight)/2-\headheight-\headsep-1in}}

\begin{document}

\title{A Multilingual FrameNet-based Grammar and Lexicon for Controlled Natural Language
\thanks{This work has been supported by the Swedish Research Council under Grant No. 2012-5746 (Reliable Multilingual Digital Communication: Methods and Applications) and by the Centre for Language Technology in Gothenburg. The research leading to these results has received funding also from the Latvian State Research Programme NexIT (Project No. 1).}}

\titlerunning{A Multilingual FrameNet-based Grammar and Lexicon for CNL}

\author{Normunds Gruzitis \and Dana Dann\'{e}lls}

\institute{N. Gruzitis \at
	University of Gothenburg, Department of Computer Science and Engineering\\
	University of Latvia, Institute of Mathematics and Computer Science\\
	\email{normunds.gruzitis@cse.gu.se, normunds.gruzitis@lu.lv}
	\and
	D. Dann\'{e}lls \at
	University of Gothenburg, Department of Swedish
}

\date{Received: date / Accepted: date} 

\maketitle

\begin{abstract}
Berkeley FrameNet is a lexico-semantic resource for English based on the theory of frame semantics. It has been exploited in a range of natural language processing applications and has inspired the development of framenets for many languages. We present a methodological approach to the extraction and generation of a computational multilingual FrameNet-based grammar and lexicon. The approach leverages FrameNet-annotated corpora to automatically extract a set of cross-lingual semantico-syntactic valence patterns. Based on data from Berkeley FrameNet and Swedish FrameNet, the proposed approach has been implemented in Grammatical Framework (GF), a categorial grammar formalism specialized for multilingual grammars. The implementation of the grammar and lexicon is supported by the design of FrameNet, providing a frame semantic abstraction layer, an interlingual semantic API (application programming interface), over the interlingual syntactic API already provided by GF Resource Grammar Library. The evaluation of the acquired grammar and lexicon shows the feasibility of the approach. Additionally, we illustrate how the FrameNet-based grammar and lexicon are exploited in two distinct multilingual controlled natural language applications. The produced resources are available under an open source license.
\end{abstract}

\section{Introduction}
\label{intro}

\citet{Kuhn2014} defines Controlled Natural Language (CNL) as ``a constructed language that is based on a certain natural language, being more restrictive concerning lexicon, syntax, and/or semantics, while preserving most of its natural properties.''
In our work, we deviate from this definition in two aspects.
First, our intention is to produce a reusable grammar that covers a restricted \emph{subset} of a natural language instead of a grammar of a predefined \emph{constructed} language.
Second, we produce a currently bilingual but potentially multilingual grammar library which is therefore not based on exactly one natural language but has a \emph{shared} semantic abstract syntax.
Thus, we do not provide a CNL as such but a high-level API (application programming interface) for the facilitation of the development of CNL grammars, making them more flexible~-- easier to modify and extend.
In a sense, we aim at bridging the gap between controlled and natural language.

\begin{figure}[b]
\centering
\includegraphics[width=1.0\columnwidth]{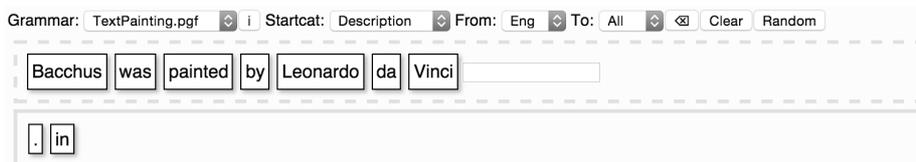}
\caption{A multilingual CNL editor developed in the MOLTO project (\url{www.molto-project.eu}). The middle row shows the sentence that has been produced so far, and the bottom row shows the choices available to the user (the possible continuation: a full stop or ``in {\textless}year{\textgreater}'').}
\label{fig:paintings}
\end{figure}

A more general aim of this research is to make existing FrameNet (FN) resources uniformly and computationally accessible for multilingual natural language generation (NLG) and controlled semantic parsing via a shared semantico-syntactic grammar and lexicon API.
We particularly consider the development of CNL interfaces to knowledge bases for authoring and verbalizing facts in a specific domain.
For example, Figure~\ref{fig:paintings} illustrates a predictive multilingual CNL editor for authoring object descriptions in the cultural heritage domain.
The detailed syntactic constructors for building the verb phrases and clauses have been manually specified for each language.
The FN-based API aims to diminish the manual efforts by providing more abstract~-- frame semantic~-- constructors, e.g. \texttt{Create\_physical\_artwork} that takes arguments \textsc{Representation} and \textsc{Creator}, and a target verb.
The future potential of our approach is to provide a means for multilingual verbalization of FN-annotated knowledge bases that have been populated by FN-based information extraction systems \citep{DasEtAl2013} and that could be automatically mapped to the appropriate frame constructors, similarly as sketched by \citet{Barzdins2014}.

At the CNL~2012 workshop, we proposed a conception of a general-purpose semantic grammar based on FrameNet~\citep{GruzitisEtAl2012}.
The proposed approach builds on the technology of Grammatical Framework (GF).
GF~\citep{Ranta2004}, a type-theoretical grammar formalism and a toolkit, offers a wide-coverage resource grammar library (RGL) for currently 30 languages which implement a shared syntactic API~\citep{Ranta2009}.
The idea behind the FrameNet-based grammar is to provide a frame semantic abstraction layer, a shared semantic API, over the syntactic API of GF RGL.

Following this conception, we successfully extracted a shared abstract syntax of wide-coverage English and Swedish grammars from FN-annotated corpora~\citep{DannellsAndGruzitis2014a}.
Soon after, we presented a more elaborated approach to automatic extraction and generation of both the shared abstract syntax and the concrete syntaxes of the proposed grammar~\citep{DannellsAndGruzitis2014b}.
In this article, we give an extended and updated presentation of the work published in the previous two papers.
Additionally, we provide the design and implementation details of FN-based lexicons for English and Swedish that are also extracted from the annotated corpora.
The experiments and tests presented here are based on the Berkeley FrameNet release 1.5 which is available as of September 2010,\footnote{\url{https://framenet.icsi.berkeley.edu/fndrupal/framenet_data}} and a snapshot of the Swedish FrameNet development version taken in December 2014.\footnote{\url{http://remu.grammaticalframework.org/framenet/SweFN_2014-12-03.zip}}

\begin{figure}[t]
\centering
\includegraphics[width=0.95\columnwidth]{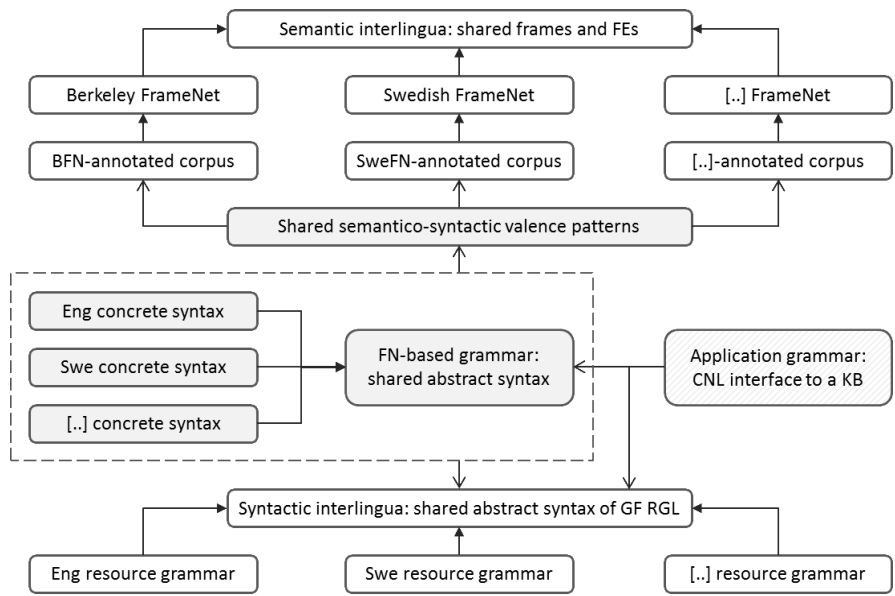}
\caption{An outline of our approach. The arrows with filled heads roughly depict the `implements'/`extends' relationship while the arrows with open heads roughly depict the `derived from'/`builds on' relationship. The grey-filled blocks are the result of our work.}
\label{fig:summary}
\end{figure}

Our approach is outlined in Figure~\ref{fig:summary}.
From the linguistic point of view, the particular characteristic is that we focus on the core argument structures (according to FrameNet), the arguments are combined compositionally, a verb is expected as the target word, and the word order is controlled according to the dominant corpus evidence.
Although we focus on English and Swedish, the same approach is intended to be applicable to other languages as well.
As a side result, we suggest a unified method for comparing and mapping semantic and syntactic valence patterns and lexical units across framenets.

The structure of this article is as follows.
In Section~\ref{sec:back}, we provide background information about FrameNet and Grammatical Framework.
Details of the FN-based approach, the experiment series and the implementation of the grammar are given in Section~\ref{sec:fn-api}.
The FN-based lexicon is further detailed in Section \ref{sec:lexicon}.
These are followed by an illustration of the use of the FN-based API in two multilingual CNL applications in Section~\ref{sec:cases}.
We provide an evaluation of our method in Section~\ref{sec:evaluation} followed by a discussion in Section~\ref{sec:future}.
Previous approaches to semantic multilingual CNL grammars are briefly discussed in Section~\ref{sec:related}.
We conclude the article in Section~\ref{sec:conclusion}.

\section{Background}
\label{sec:back}

\subsection{Berkeley FrameNet (BFN)}
\label{ssec:fn}

Berkeley FrameNet \citep{FillmoreEtAl2003} is a lexico-semantic
resource based on the theory of frame semantics
\citep{Fillmore1985}.\footnote{\url{https://framenet.icsi.berkeley.edu/}}
According to this theory, a semantic \emph{frame} representing a
cognitive scenario is characterized in terms of \emph{frame
  elements}~(FE) and is evoked by target words called \emph{lexical
  units}~(LU).

FEs are classified in four groups: core, core unexpressed, peripheral
and extra-thematic \citep{RuppenhoferEtAl2010}.  A set of core FE
instantiates the conceptually necessary components of a frame, and
uniquely characterizes the frame, making it different from other
frames. Core unexpressed FEs are core FEs that may not be used in
descendant frames. FEs, such as \textsc{Time}, \textsc{Place},
\textsc{Manner}, that do not uniquely characterize a frame, and can be
instantiated in any semantically appropriate frame are classified as
peripheral. Extra-thematic FEs do not have a frame-specific
understanding, unlike core and peripheral FEs.

In our work, we distinguish two classes of FEs: \emph{core} that
includes core unexpressed, and \emph{non-core} that includes
peripheral and extra-thematic.  Core FEs syntactically tend to
correspond to verb arguments, in contrast to non-core FEs that
typically are adjuncts.

As an example, consider the frame \texttt{Desiring} given in Table~\ref{tab:FrameEng} where we find:
\begin{inparaenum}[(i)]
\item the definition of the frame, a lexicographic description of the scenario it represents, and
\item lists of core and non-core FEs, the semantic roles.
\end{inparaenum}

\begin{table}[b]
\tabcolsep 4pt
\begin{center}
\begin{tabular}{|l|l|}
\hline
\multicolumn{2}{|c|}{\texttt{Desiring}}\\
\hline
Definition:
& An {\sc Experiencer} desires that an {\sc Event} occur. In some cases, the\\
& {\sc Experiencer} is an active participant in the {\sc Event}, and in such\\
& cases the {\sc Event} itself is often not mentioned, but rather some \\
& {\sc Focal\_participant} which is subordinately involved.\\
\hline
Core FEs:
& {\sc Event}, {\sc Experiencer}, {\sc Focal\_participant}, {\sc Location\_of\_Event}\\
\hline
Non-core FEs:
& {\sc Cause}, {\sc Degree}, {\sc Duration}, {\sc Manner}, {\sc Place}, {\sc Purpose\_of\_Event},\\
&  {\sc Reason}, {\sc Role\_of\_focal\_participant}, {\sc Time}, {\sc Time\_of\_Event}\\
\hline
\end{tabular}
\caption{The definition and FE groups of the FrameNet frame \texttt{Desiring}.}
\label{tab:FrameEng}
\end{center}
\end{table}

An LU entry is a pairing of a lemma and a frame, and it carries both
the semantic and the syntactic valence information about the possible
realizations of the FEs participating in the frame. The syntactic
valence is language-specific, and the valence patterns are derived
from FN-annotated corpora.  The syntactic constituents of the example
sentences are annotated according to COMLEX Syntax
\citep{MeyersEtAl95}.  To take an example, consider the valence
patterns for the verb \emph{want} given in Table~\ref{tab:feBFN}.

\begin{table}[t]
\tabcolsep 4pt
\begin{center}
\begin{tabular}{|c|ll|}
\hline
Examples & \multicolumn{2}{|l|}{Valence patterns}\\
\hline
40 & {\sc Event} & {\sc Experiencer}\\
(22) & $\mathsf{VPto.Dep}$ & $\mathsf{NP.Ext}$\\
\hline
14 & {\sc Experiencer} & {\sc Focal\_participant}\\
(10) & $\mathsf{NP.Ext}$ & $\mathsf{NP.Obj}$\\
(1) & $\mathsf{PP[}$$\mathit{by}$$\mathsf{].Dep}$ & $\mathsf{NP.Ext}$\\
\hline
\end{tabular}
\caption{Some semantic patterns and some of their syntactic
  realizations found in BFN for the LU \emph{want.v} evoking
  the frame \texttt{Desiring}. The syntactic annotations include phrase types,
  e.g. noun phrase ($\mathsf{NP}$), prepositional phrase
  ($\mathsf{PP}$), verb phrase ($\mathsf{VP}$), and shallow
  grammatical functions: external argument ($\mathsf{Ext}$), first
  object ($\mathsf{Obj}$), general dependent ($\mathsf{Dep}$).}
\label{tab:feBFN}
\end{center}
\end{table}

There have been also experiments on automatic alignment of LUs in BFN
to synsets in Princeton WordNet \citep{FerrandezEtAl2010}, a
complementary resource that would help to extend the coverage of BFN,
and link LUs across languages.  These links, however, are not
available as a part of the FrameNet data release.

The FrameNet approach to frame semantics, approbated in BFN, provides
a benchmark for representing large amounts of word senses and word
usage patterns through the linguistic annotation of corpus examples,
therefore the exploitation of FN-like resources has been appealing for
a range of advanced NLP applications such as semantic
parsing~\citep{DasEtAl2013}, information
extraction~\citep{MoschittiEtAl2003} and natural language
generation~\citep{RothAndFrank2009}.  There are FNs available for
German, Japanese, Spanish~\citep{Boas2009} and
Swedish~\citep{BorinEtAl2010}.  More initiatives exist for other
languages.  In this article, we consider Berkeley FrameNet and Swedish
FrameNet.

\subsection{Swedish FrameNet (SweFN)}
\label{ssec:Swefn}

Swedish FrameNet has been developed within the Swedish
FrameNet++ project at Språkbanken
\citep{BorinEtAl2010}.\footnote{\url{http://spraakbanken.gu.se/swefn/}}
One of the aims of the project was to integrate a number of existing
Swedish lexical resources and harmonize the information they
contain. All the integrated resources are linked to the lexical
entries of SALDO \citep{BorinEtAl2013}, an association lexicon which
contains morphological and lexical-semantic information for more than
125,000 Swedish words of which 13,000 are
verbs.\footnote{\url{http://spraakbanken.gu.se/saldo}} SALDO is
therefore considered as the pivot of these integrated Swedish lexical
resources.

Integrated lexical resources linked to one pivot lexicon have the
advantage of yielding large amount of information. Some of the
information we gain access to is syntactic valence information for
verbs from SIMPLE and PAROLE lexicons \citep{LenciEtAl2000}, and synsets
and senses from WordNet.\footnote{\url{http://spraakbanken.gu.se/swe/resurs/wordnet-saldo}}

The SweFN resource has been expanded from BFN, which means it follows
the same structure and theoretical principles that have been taken in
BFN. For example, the description of the frame \texttt{Desiring} shown
in Table~\ref{tab:FrameEng} is the same also in SweFN. This is the
language independent aspect. Regarding the language dependent aspect,
there are other, more practical differences between the
resources. These differences concern the content of the frames including: lexical units from SALDO,
compound pattern analysis and examples, syntactic annotations of the
example sentences, and domain specification of frames.

Some of the LUs, e.g. the SALDO entries, that evoke the frame
\texttt{Desiring} are: \emph{känna\_för.vb.1} `to feel like',
\emph{längta.vb.1} `to yearn', \emph{vilja.vb.1} `to want',
\emph{åtrå.vb.1} `to desire', \emph{begärelse.nn.1} `wish',
\emph{åtrå.nn.2} `desire'. The number indicates SALDO's sense
identifier, `vb' and `nn' indicate the part-of-speech (POS) tags.

All example sentences in SweFN are syntactically annotated with the Swedish version of
MaltParser for depedency structures \citep{NivreEtAl2004}. The
annotation scheme is based on KORP, Språkbanken's corpus
infrastructure, part-of-speech and morphosyntactic markup tag sets
\citep{BorinEtAl2012}.\footnote{\url{http://spraakbanken.gu.se/eng/korp-info}}
Table~\ref{tab:feSweFN} shows some semantic valence patterns and their
syntactic realizations for the verb \emph{känna för} `to feel like'.

\begin{table}
\tabcolsep 4pt
\begin{center}
\begin{tabular}{|c|ll|}
\hline
Examples & \multicolumn{2}{|l|}{Valence patterns}\\
\hline
1 & {\sc Event} & {\sc Experiencer}\\
(1) & $\mathsf{VB.INF.VG}$ & $\mathsf{NN.SS}$\\
\hline
2 & {\sc Experiencer} & {\sc Focal\_participant}\\
(2) & $\mathsf{NN.SS}$ & $\mathsf{NN.OO}$\\
\hline
\end{tabular}
\caption{Semantic patterns and their syntactic realizations found in
  SweFN for \emph{känna\_för.vb} evoking \texttt{Desiring}. The
  syntactic patterns include morpho-syntactic tags, e.g. noun
  ($\mathsf{NN}$), verb ($\mathsf{VB}$), infinite form
  ($\mathsf{INF}$), and dependency labels, e.g. subject ($\mathsf{SS}$), direct object ($\mathsf{OO}$).}
\label{tab:feSweFN}
\end{center}
\end{table}

As we mentioned, SweFN mostly uses the BFN frame inventory, however,
around 50 additional frames are introduced in SweFN, and around
15 BFN frames have been modified. What characterizes the modified
frames is change of core FEs and change of the
frame content that is either more specific or less specific.

\subsection{Grammatical Framework (GF)}
\label{ssec:gf}

The presented grammar is implemented in GF, a categorial grammar
formalism specialized for multilingual (parallel)
grammars~\citep{Ranta2004}. One of the key features of GF grammars is
the separation between an abstract syntax and concrete syntaxes. The
abstract syntax defines the language-independent structure, the
semantics of a domain-specific application grammar or a
general-purpose grammar library, while the concrete syntaxes define
the language-specific syntactic and lexical realization of the
abstract syntax.

Remarkably, GF is not only a grammar formalism or a programming
language. It also provides a reusable general-purpose resource grammar library
(RGL) for currently 30 languages that implement the same abstract syntax, a
shared syntactic API~\citep{Ranta2009}. The grammars implement common
syntactic constructions and describe the inflectional morphology of a
language. The use of the shared syntactic types and functions allows
for rapid and rather flexible development of multilingual application
grammars without the need of specifying low-level details like
inflectional paradigms, syntactic agreement and word order.
In order to hide the low-level details, RGL has a high-level interface that provides constructors like $\mathsf{mkCl: NP \to VP \to Cl}$ for building a clause from a NP and a VP.\footnote{\url{http://www.grammaticalframework.org/lib/doc/synopsis.html}}

\section{FrameNet-Based Grammar}
\label{sec:fn-api}

The language-independent layer of FrameNet, i.e. frames and FEs~--
the semantic valence~-- is defined in the abstract syntax of the
proposed multilingual grammar library, while the language-specific
layers, i.e. the surface realization of frames and LUs~-- the syntactic
valence~-- is defined in concrete
syntaxes.\footnote{\url{https://github.com/GrammaticalFramework/gf-contrib/tree/master/framenet}\\(The
  acquired grammar and lexicon; version 0.9.7 at the time of
  writing.)} The syntactic API of GF RGL is used for generalizing and
unifying the grammatical types and constructions used in different
framenets, which facilitates porting the implementation to other
languages. The FrameNet-based grammar, in turn, provides a frame
semantic abstraction layer to RGL, so that the application grammar
developer can primarily manipulate with plain semantic constructors in
combination with some simple syntactic constructors instead of
comparatively complex syntactic constructors for building verb phrases
(VP). Moreover, the frame constructors can be typically specified for
all languages at once in the shared concrete syntax (functor) of an
application grammar.

In this research, we consider only those frames for which there is at
least one corpus example where the frame is evoked by a verb. In
addition, we consider only core FEs which uniquely characterize the
frame.

BFN version 1.5 defines 1,020 frames, of which, according to our
calculations, 559 are evoked by 3,254 verb LUs in 69,260 annotated
sentences. As of December 2014, the SweFN development version covers
995 frames of which 660 are evoked by 2,887 verb LUs in 4,400
annotated sentences.

\subsection{Abstract Syntax}
\label{ssec:abstract}

To acquire a common abstract syntax, a common semantic API,\footnote{\url{http://www.grammaticalframework.org/framenet/}} we have extracted a set of shared semantico-syntactic frame valence patterns from the annotated sentences in BFN and SweFN. For instance, the shared valence patterns for the frame $\mathsf{Desiring}$ are:

\begin{enumerate}
\item[] $\mathsf{Desiring/V_{Act}}$ $\mathsf{Experiencer/NP_{nsubj}}$ $\mathsf{Focal\_participant/Adv}$
\item[] $\mathsf{Desiring/V2_{Act}}$ $\mathsf{Experiencer/NP_{nsubj}}$ $\mathsf{Focal\_participant/NP_{dobj}}$
\item[] $\mathsf{Desiring/VV_{Act}}$ $\mathsf{Event/VP}$ $\mathsf{Experiencer/NP_{nsubj}}$
\end{enumerate}

\noindent which correspond, for instance, to the following annotated examples in BFN:

\begin{enumerate}
\item[] {[\emph{Dexter}]}\textsubscript{$\mathsf{Experiencer/NP}$} [\emph{YEARNED}]\textsubscript{$\mathsf{V}$} [\emph{for a cigarette}]\textsubscript{$\mathsf{Focal\_participant/Adv}$}
\item[] {[\emph{she}]}\textsubscript{$\mathsf{Experiencer/NP}$} [\emph{WANTS}]\textsubscript{$\mathsf{V2}$} [\emph{a protector}]\textsubscript{$\mathsf{Focal\_participant/NP}$}
\item[] {[\emph{I}]}\textsubscript{$\mathsf{Experiencer/NP}$} \emph{would n't} [\emph{WANT}]\textsubscript{$\mathsf{VV}$} [\emph{to know}]\textsubscript{$\mathsf{Event/VP}$}
\end{enumerate}

The actual BFN phrase types ($\mathsf{NP}$, various subtypes of $\mathsf{VP}$, $\mathsf{PP}$ and its subtypes, etc.) are generalized to the RGL types $\mathsf{NP}$, $\mathsf{VP}$, $\mathsf{Adv}$, $\mathsf{S}$ and $\mathsf{QS}$.\footnote{Where $\mathsf{Adv}$ is a VP-modifying adverb, $\mathsf{S}$~-- an embedded declarative sentence, and $\mathsf{QS}$~-- an embedded question.}
Verb types ($\mathsf{V}$, $\mathsf{V2}$, $\mathsf{V3}$, $\mathsf{VV}$, $\mathsf{VS}$, $\mathsf{VQ}$, $\mathsf{V2V}$, $\mathsf{V2S}$ and $\mathsf{V2Q}$\footnote{Where $\mathsf{V}$ is a one-place verb, $\mathsf{V2}$~-- a two-place verb, $\mathsf{V3}$~-- a three-place verb, $\mathsf{VV}$~-- a $\mathsf{VP}$-complement verb, $\mathsf{VS}$~-- an $\mathsf{S}$-complement verb, $\mathsf{VQ}$~-- a $\mathsf{QS}$-complement verb, $\mathsf{V2V}$~-- a verb with $\mathsf{NP}$ and $\mathsf{VP}$ complements, $\mathsf{V2S}$~-- a verb with $\mathsf{NP}$ and $\mathsf{S}$ complements, and $\mathsf{V2Q}$~-- a verb with $\mathsf{NP}$ and $\mathsf{QS}$ complements.}) are inferred from the syntactic valence of LUs in the respective examples.

In addition to phrase types, the extracted valence patterns also specify inferred grammatical relations of $\mathsf{NP}$-typed FEs: $\mathsf{nsubj}$ (subject), $\mathsf{nsubjpass}$ (passive subject), $\mathsf{dobj}$ (direct object) and $\mathsf{iobj}$ (indirect object) that correspond to the universal dependency relations \citep{MarneffeEtAl2014}.
Therefore, we also include the grammatical voice  ($\mathsf{Act}$/$\mathsf{Pass}$\footnote{\url{http://universaldependencies.github.io/docs/u/feat/Voice.html}}) in the pattern comparison and in pattern identifiers used in the abstract syntax. It is not necessary, however, to reflect the grammatical relations in the abstract syntax; this knowledge is taken into account in the pattern comparison and while generating the concrete syntaxes, but it is not required in order to use the resulting grammar as an API.

\subsubsection{Sentence patterns versus normalized valence patterns}
\label{sssec:patterns}

The first step in the extraction of shared valence patterns is to convert the annotated corpus examples into more general and uniform sentence patterns~-- valence patterns that preserve the word-order (the order of FEs in a particular sentence), subcategorize FEs by the grammatical RGL types and include the universal  grammatical relations of $\mathsf{NP}$-typed FEs (both inferred from the particular sentence), include prepositions or cases that are used to realize $\mathsf{Adv}$-typed FEs (for deciding on LU-specific or even frame-specific default values in the future), and include references to the target verbs (LUs). Duplicate sentence patterns are kept in the output for frequency counts.

The conversion process is language- and framenet-specific because there is no unified annotation model used across framenets. BFN and SweFN use not only different XML schemas and POS tagsets; they also use different approaches for annotating the syntactic structure of a sentence.

In BFN, a phrase-structure approach is used, which is complemented by few shallow grammatical relations: an external argument (a phrase outside the VP of the target verb), the first object in the active voice (either direct or indirect), and a general dependent. A simplified excerpt from the BFN corpus for the verb \emph{want} evoking the frame \texttt{Desiring} is:

\begingroup
\fontsize{8pt}{10pt}\selectfont
\begin{Verbatim}[xleftmargin=5.5mm,commandchars=\\\{\}]
<sentence>
  <text>\emph{Traders in the city want a change.}</text>
  <annotationSet>
    <layer rank="1" name="BNC">
      <label start="0" end="6" name="NP0"/>
      <label start="20" end="23" name="VVB"/>
      <label start="25" end="25" name="AT0"/>
    </layer>
  </annotationSet>
  <annotationSet status="MANUAL">
    <layer rank="1" name="FE">
      <label start="0" end="18" name="Experiencer"/>
      <label start="25" end="32" name="Event"/>
    </layer>
    <layer rank="1" name="GF">
      <label start="0" end="18" name="Ext"/>
      <label start="25" end="32" name="Obj"/>
    </layer>
    <layer rank="1" name="PT">
      <label start="0" end="18" name="NP"/>
      <label start="25" end="32" name="NP"/>
    </layer>
    <layer rank="1" name="Target">
      <label start="20" end="23" name="Target"/>
    </layer>
  </annotationSet>
</sentence>
\end{Verbatim}
\endgroup

In SweFN, a dependency approach is used. A simplified excerpt from the SweFN corpus for the verb \emph{vilja} `want' evoking the frame \texttt{Desiring} is:\footnote{SweFN tags are described at \url{http://stp.lingfil.uu.se/~nivre/swedish_treebank/}}

\begingroup
\fontsize{8pt}{10pt}\selectfont
\begin{Verbatim}[xleftmargin=5.5mm,commandchars=\\\{\}]
<sentence>
  <w pos="JJ" ref="1" dephead="2" deprel="DT">\emph{Nästa}</w>
  <w pos="NN" ref="2" dephead="3" deprel="TA">\emph{gång}</w>
  <w pos="VB" ref="3" deprel="ROOT">\emph{skulle}</w>
  <element name="Experiencer">
    <w pos="PN" ref="4" dephead="3" deprel="SS">\emph{jag}</w>
  </element>
  <element name="LU">
    <w msd="VB.AKT" ref="5" dephead="3" deprel="VG">\emph{vilja}</w>
  </element>
  <element name="Event">
    <w msd="VB.INF" ref="6" dephead="5" deprel="VG">\emph{ha}</w>
    <w pos="RG" ref="7" dephead="8" deprel="DT">\emph{sju}</w>
    <w pos="NN" ref="8" dephead="6" deprel="OO">\emph{sångare}</w>
  </element>
</sentence>
\end{Verbatim}
\endgroup

It should be noted that a characteristic of BFN is that FEs which are
missing in the sentence are still annotated if the grammar allows or
requires the omission, or the identity/type of an FE is understood
from the context \citep{RuppenhoferEtAl2010}. Such FEs would be
potentially interesting to consider, however, we ignore them as they
have no grammatical annotations.

Because of the partial and often erroneous grammatical annotations,
various framenet-specific rules and heuristics are applied for
generalizing to RGL types, for inferring the grammatical voice and
relations, and for partially correcting the automatic annotation
errors.

When the uniform sentence patterns are acquired for all languages, a
common language- and framenet-independent processor is used in all
the remaining steps, including the generation of the abstract and
concrete syntaxes and lexicons.

Sentence patterns are summarized and grouped into normalized valence
patterns ignoring the word order and prepositions (or cases). As an
example, a partial summary of patterns for the frame \texttt{Desiring}
in BFN is:

\begingroup
\fontsize{8pt}{10pt}\selectfont
\begin{Verbatim}[xleftmargin=5.5mm,commandchars=\\\{\}]
Act : 275
  Event/VP Experiencer/NP.nsubj : 61
    Experiencer/NP.nsubj Event/VP : 59
    Event/VP Experiencer/NP.nsubj : 2
  Experiencer/NP.nsubj Focal_participant/NP.dobj : 61
    Experiencer/NP.nsubj Focal_participant/NP.dobj : 55
    Focal_participant/NP.dobj Experiencer/NP.nsubj : 6
  Experiencer/NP.nsubj Focal_participant/Adv : 43
    Experiencer/NP.nsubj Focal_participant/Adv[\emph{for}] : 26
    Experiencer/NP.nsubj Focal_participant/Adv[\emph{after}] : 7
    Experiencer/NP.nsubj Focal_participant/Adv : 2
    ...
  ...
Pass : 13
  Experiencer/NP.dobj Focal_participant/NP.nsubjpass : 5
    Focal_participant/NP.nsubjpass Experiencer/NP.dobj : 5
  ...
\end{Verbatim}
\endgroup

For generating the abstract syntax, we consider only the normalized valence patterns. The most frequent sentence pattern of each normalized pattern contains sufficient information for generating the concrete syntax for the respective language.

\subsubsection{Experiment series}
\label{sssec:series}

To roughly estimate the impact of certain decisions that have been made in the automatic extraction of the semantico-syntactic valence patterns, we have run a series of experiments with various settings:

\begin{enumerate}[leftmargin=1.25cm]
\item[\texttt{0.0}] Extract sentence patterns using the framenet-specific grammatical types (baseline).
\item[\texttt{1.0}] In addition to 0.0, skip examples containing currently unconsidered realizations of FEs, namely quotation and few subtypes of $\mathsf{S}$ (3.4\% of BFN examples; no SweFN examples).\footnote{Additionally, more than 100 examples are skipped in both corpora due to inconsistent semantico-syntactic annotations that were not fixed by the current heuristics.}
\item[\texttt{2.0}] In addition to 1.0, generalize the grammatical types according to GF RGL.
\item[\texttt{3.0}] In addition to 2.0, skip once-used valence patterns (if the frame has at least one pattern that is used more than once).
\end{enumerate}

\noindent where each series include two subseries:

\begin{enumerate}[leftmargin=1.25cm]
\item[\texttt{x.A}] Skip repeated FEs (mostly due to coordination, wh-words making discontinuous PPs, and anchors of relative clauses).\footnote{If repeated FEs are of different RGL types, the whole example is currently skipped.}
\item[\texttt{x.B}] Skip non-core FEs and repeated FEs.
\end{enumerate}

The results are summarized in Tables~\ref{bfn_statistics} and~\ref{swefn_statistics}. We are primarily interested in Settings~2.B and 3.B which seem to be optimal for SweFN and BFN respectively: the number of covered frames slightly decreases, but it makes the resulting patterns more prototypical and significantly reduces the potential number of API functions to be generated in the abstract syntax. For a large corpus like BFN, skipping once-used valence patterns helps to reduce the propagation of annotation errors, but, for a relatively small corpus like SweFN, it is not reasonable. In the future, it would be reasonable, however, to include typical non-core FEs in the resulting valence patterns: this would slightly increase the average number of FEs from 2 to 3.

\begin{table}[t]
\begin{center}
\begin{tabular}{|c|c|c|rr|rr|rr|}
\hline
\multicolumn{1}{|c}{\multirow{2}{*}{\rotatebox[origin=c]{90}{Settings~}}} & \multicolumn{1}{|c}{\multirow{2}{*}{\rotatebox[origin=c]{90}{Frames~~}}} & \multicolumn{1}{|c}{\multirow{2}{*}{\rotatebox[origin=c]{90}{LUs~~~~}}} & \multicolumn{2}{|c}{\begin{tabular}[c]{@{}c@{}}Valence\\patterns\end{tabular}} & \multicolumn{2}{|c}{\begin{tabular}[c]{@{}c@{}}Sentence\\patterns\end{tabular}} & \multicolumn{2}{|c|}{\begin{tabular}[c]{@{}c@{}}Corpus\\examples\end{tabular}}\\
\cline{4-9}
\multicolumn{1}{|c}{} & \multicolumn{1}{|c}{} & \multicolumn{1}{|c}{} & \multicolumn{1}{|c}{total} & \multicolumn{1}{|c}{\begin{tabular}[c]{@{}c@{}}per\\frame\end{tabular}} & \multicolumn{1}{|c}{total} & \multicolumn{1}{|c}{\begin{tabular}[c]{@{}c@{}}{\small per}\\{\small valence}\\{\small pattern}\end{tabular}} & \multicolumn{1}{|c}{total} & \multicolumn{1}{|c|}{\begin{tabular}[c]{@{}c@{}}{\small per}\\{\small sentence}\\{\small pattern}\end{tabular}}\\
\hline
{\tt 0.0} & 559 & 3254 & 20067 & 36 & 25905 & 1.3 & 69260 & 2.7\\
\hline
{\tt 2.0} & 558 & 3237 & 16564 & 30 & 24642 & 1.5 & 66918 & 2.7\\
{\tt 2.A} & 554 & 3232 & 14202 & 26 & 22256 & 1.6 & 65575 & 2.9\\
{\tt 2.B} & 554 & 3232 & 5489 & 10 & 8719 & 1.6 & 65670 & 7.5\\
\hline
{\tt 3.0} & 558 & 3237 & 6810 & 12 & 14888 & 2.2 & 57164 & 3.8\\
{\tt 3.A} & 554 & 3232 & 6355 & 11 & 14409 & 2.3 & 57728 & 4.0\\
{\tt 3.B} & 554 & 3232 & 3666 & 7 & 6896 & 1.9 & 63847 & 9.3\\
\hline
\end{tabular}
\caption{Experiment series for extracting semantico-syntactic valence patterns from BFN. Sentence patterns preserve the order of FEs and prepositions/cases of $\mathsf{Adv}$-typed FEs. Normalized valence patterns disregard both.}
\label{bfn_statistics}
\end{center}
\end{table}

\begin{table}[t]
\begin{center}
\begin{tabular}{|c|c|c|rr|rr|rr|}
\hline
\multicolumn{1}{|c}{\multirow{2}{*}{\rotatebox[origin=c]{90}{Settings~}}} & \multicolumn{1}{|c}{\multirow{2}{*}{\rotatebox[origin=c]{90}{Frames~~}}} & \multicolumn{1}{|c}{\multirow{2}{*}{\rotatebox[origin=c]{90}{LUs~~~~}}} & \multicolumn{2}{|c}{\begin{tabular}[c]{@{}c@{}}Valence\\patterns\end{tabular}} & \multicolumn{2}{|c}{\begin{tabular}[c]{@{}c@{}}Sentence\\patterns\end{tabular}} & \multicolumn{2}{|c|}{\begin{tabular}[c]{@{}c@{}}Corpus\\examples\end{tabular}}\\
\cline{4-9}
\multicolumn{1}{|c}{} & \multicolumn{1}{|c}{} & \multicolumn{1}{|c}{} & \multicolumn{1}{|c}{total} & \multicolumn{1}{|c}{\begin{tabular}[c]{@{}c@{}}per\\frame\end{tabular}} & \multicolumn{1}{|c}{total} & \multicolumn{1}{|c}{\begin{tabular}[c]{@{}c@{}}{\small per}\\{\small valence}\\{\small pattern}\end{tabular}} & \multicolumn{1}{|c}{total} & \multicolumn{1}{|c|}{\begin{tabular}[c]{@{}c@{}}{\small per}\\{\small sentence}\\{\small pattern}\end{tabular}}\\
\hline
{\tt 0.0} & 660 & 2887 & 4069 & 6 & 4111 & 1.0 & 4400 & 1.1\\
\hline
{\tt 2.0} & 658 & 2834 & 3388 & 5 & 3578 & 1.1 & 4267 & 1.2\\
{\tt 2.A} & 654 & 2828 & 3300 & 5 & 3495 & 1.1 & 4180 & 1.2\\
{\tt 2.B} & 654 & 2828 & 2255 & 3 & 2432 & 1.1 & 4191 & 1.7\\
\hline
{\tt 3.0} & 658 & 2834 & 1975 & 3 & 2165 & 1.1 & 2854 & 1.3\\
{\tt 3.A} & 654 & 2828 & 1950 & 3 & 2145 & 1.1 & 2830 & 1.3\\
{\tt 3.B} & 654 & 2828 & 1401 & 2 & 1578 & 1.1 & 3337 & 2.1\\
\hline
\end{tabular}
\caption{Experiment series for extracting semantico-syntactic valence patterns from SweFN.}
\label{swefn_statistics}
\end{center}
\end{table}

\subsubsection{Shared valence patterns}
\label{sssec:shared}

The extracted sets of semantico-syntactic valence patterns can vary across languages depending on corpora. Having multilingual applications in mind, we are primarily interested in valence patterns whose implementation can be generated, based on corpus evidence, for all considered languages. Thus, we focus on valence patterns that are shared between framenets. The multilingual criterion also helps in reducing the number of incorrectly derived patterns due to annotation errors introduced by the automatic POS tagging and syntactic parsing applied in both BFN and SweFN corpora. Frequent patterns that are not verified across framenets could be separated into language-specific extra modules of the library (in a similar way as it is done with some language-specific syntactic phenomena in RGL).

To find a representative yet condensed set of shared valence patterns, we compare the extracted normalized patterns by subsumption instead of equivalence. Pattern $\mathsf{A}$ subsumes pattern $\mathsf{B}$ if:

\begin{enumerate}[leftmargin=1cm]
\item $\mathsf{A.frame = B.frame}$
\item $\mathsf{A.verbType = B.verbType}$
\item $\mathsf{A.grammaticalVoice = B.grammaticalVoice}$
\item $\mathsf{B.FEs \subseteq A.FEs}$\footnote{Taking into account the grammatical types and relations.}
\end{enumerate}

If $\mathsf{A}$ subsumes $\mathsf{B}$ and $\mathsf{B}$ subsumes $\mathsf{A}$ then $\mathsf{A}$ equals $\mathsf{B}$. If a pattern of $\mathsf{FN_1}$ is subsumed by a pattern of $\mathsf{FN_2}$, it is added to the shared set (and vice versa). In the final shared set, patterns which are subsumed by other patterns in the set are removed. For instance, in the following example, $\mathsf{P1}$ is subsumed by $\mathsf{P2}$, $\mathsf{P3}$ is subsumed by $\mathsf{P1}$ and $\mathsf{P2}$, $\mathsf{P1}$ and $\mathsf{P3}$ are to be removed:

\begin{enumerate}
\item[] $\mathsf{P1}$: $\mathsf{Apply\_heat/V2_{Act}}$ $\mathsf{Cook/NP_{nsubj}}$ $\mathsf{Food/NP_{dobj}}$
\item[] $\mathsf{P2}$: $\mathsf{Apply\_heat/V2_{Act}}$ $\mathsf{Container/Adv}$ $\mathsf{Cook/NP_{nsubj}}$ $\mathsf{Food/NP_{dobj}}$
\item[] $\mathsf{P3}$: $\mathsf{Apply\_heat/V2_{Act}}$ $\mathsf{Food/NP_{dobj}}$
\end{enumerate}

This approach is supported by the design of the FrameNet-based grammar which accepts an empty phrase as an argument to a frame building function if the corresponding FE is not expressed in the sentence.

The comparison is first done between BFN and SweFN sets of verb frames (Table~\ref{compare-frames}) and then between sets of valence patterns that belong to the shared set (intersection) of verb frames (Table~\ref{compare-patterns}). For a number of shared frames, no shared valence patterns are found, therefore the final set of shared frames is smaller (Table~\ref{compare-patterns}). Intuitively, this is partly because of the size of SweFN (about 6 examples per frame in SweFN versus more than 115 examples per frame in BFN) and partly because of non-compositionality across languages.

\begin{table*}[b]
\tabcolsep 3pt
\begin{center}
\begin{tabular}{|c|c|c|rr|rr|c|rr|}
\hline
\multicolumn{1}{|c}{\multirow{2}{*}{\begin{tabular}[c]{@{}c@{}}Settings\\Eng:Swe\end{tabular}}} & \multicolumn{9}{|c|}{Intermediate sets of verb frames}\\
\cline{2-10}
\multicolumn{1}{|c}{} & \multicolumn{1}{|c}{Eng} & \multicolumn{1}{|c}{Swe} & \multicolumn{2}{|c}{Eng$\setminus$Swe} & \multicolumn{2}{|c}{Swe$\setminus$Eng} & \multicolumn{1}{|c}{Eng$\cup$Swe} & \multicolumn{2}{|c|}{Eng$\cap$Swe}\\
\hline
{\tt 2.B/3.B:2.B} & 554 & 654 & 31 & (6\%) & 131 & (20\%) & 685 & 523 & (76\%)\\
\hline
\end{tabular}
\caption{Comparison of verb frames found in BFN and SweFN. Symbols $\setminus$, $\cup$ and $\cap$ denote the set operations \emph{difference}, \emph{union} and \emph{intersection}.}
\label{compare-frames}
\end{center}
\end{table*}

\begin{table*}[b]
\tabcolsep 3pt
\begin{center}
\begin{tabular}{|c|r|r|rr|rr|c|rr|c|c|}
\hline
\multicolumn{1}{|c}{\multirow{2}{*}{\begin{tabular}[c]{@{}c@{}}Settings\\Eng:Swe\end{tabular}}} & \multicolumn{9}{|c|}{Intermediate sets of valence patterns} & \multicolumn{2}{|c|}{Final sets}\\
\cline{2-12}
\multicolumn{1}{|c}{} & \multicolumn{1}{|c}{Eng} & \multicolumn{1}{|c}{Swe} & \multicolumn{2}{|c}{Eng$\setminus$Swe} & \multicolumn{2}{|c}{Swe$\setminus$Eng} & \multicolumn{1}{|c}{Eng$\cup$Swe} & \multicolumn{2}{|c|}{Eng$\cap$Swe} & Patterns & Frames\\
\hline
{\tt 2.B:2.B} & 4546 & 1854 & 2920 & (64\%) & 643 & (35\%) & 5289 & 1726 & (33\%) & 944 & 489\\
{\tt 3.B:2.B} & 3020 & 1854 & 1669 & (55\%) & 745 & (40\%) & 3879 & 1465 & (38\%) & 869 & 483\\
\hline
\end{tabular}
\caption{Comparison of valence patterns of the shared frames found in BFN and SweFN.}
\label{compare-patterns}
\end{center}
\end{table*}

In the result, from around 64,000 annotated sentences in BFN (Settings 3.B) and around 4,200 annotated sentences in SweFN (Settings 2.B), we have extracted a set of 869 shared valence patterns covering 483 frames. The result is a proper subset of what would be acquired if Settings 2.B were applied to BFN.

The 869 valence patterns reuse 541 semantico-syntactic types: 339 FEs of type $\mathsf{NP}$, 159 FEs of type $\mathsf{Adv}$, 17 FEs of type $\mathsf{VP}$, 17 FEs of type $\mathsf{S}$ and 9 FEs of type $\mathsf{QS}$. If considering only the semantic types, there are 429 different FEs.


\subsubsection{Implementation}
\label{sssec:implement}

The shared valence patterns are declared as frame building functions (henceforth called frame functions) that take one or more core FEs and one target verb as arguments. FEs are expected in the alphabetical order while the verb is always the last argument. The language-specific word order is specified in the concrete syntaxes.

For each frame, the set of core FEs is often split into several alternative functions according to the corpus evidence.\footnote{It is often practically impossible or uncommon that all core FEs are used in the same sentence. For instance, {\sc Area} is mutually exclusive with five other core FEs in the frame \texttt{Motion}, and these five other $\mathsf{Adv}$-typed FEs normally are not used altogether.} Different subsets of core FEs may require different types of target verbs. We also differentiate between functions that return clauses in the passive voice from functions that return active voice clauses because the subject and object FEs swap their grammatical relations and/or the order\footnote{E.g. in a highly inflected language.} that is not reflected in the abstract syntax.

The verb type is always added as a suffix to the function name, and the voice tag is appended in the case of the passive voice. If this is not sufficient to make the function name unique, a discriminative number is appended as well. For instance, consider the following abstract functions derived from the extracted valence patterns given at the beginning of Section~\ref{ssec:abstract}:

\begingroup
\fontsize{9.35pt}{11.35pt}\selectfont
\begin{enumerate}
\item[] $\mathit{fun}$ $\mathsf{Desiring\_V : Experiencer\_NP \to Focal\_participant\_Adv \to V \to Clause}$
\item[] $\mathit{fun}$ $\mathsf{Desiring\_VV : Event\_VP \to Experiencer\_NP \to VV \to Clause}$
\item[] $\mathit{fun}$ $\mathsf{Desiring\_V2 : Experiencer\_NP \to Focal\_participant\_NP \to V2 \to Clause}$
\item[] $\mathit{fun}$ $\mathsf{Desiring\_V2\_Pass : Experiencer\_NP \to Focal\_participant\_NP \to V2 \to Clause}$\footnote{$\mathsf{Desiring\_V2\_Pass}$ is included for illustration but is not directly acquired from a shared pattern. Missing passive or active voice patterns could be acquired implicitly~-- deriving them from the corresponding active or passive voice patterns. However, for now we are strictly following the corpus evidence.}
\end{enumerate}
\endgroup

In GF, constituents and features of phrases are stored in objects of record types, and functions are applied to such objects to construct phrase trees. In the abstract syntax, both argument types and the value type of a function are separated by right associative arrows, i.e. all functions are curried. Arguments of a frame function are combined into an object of type $\mathsf{Clause}$ that differs from the RGL type $\mathsf{Cl}$. A $\mathsf{Clause}$ whose linearization type is $\mathsf{\{np : NP ; vp : VP\}}$ comprises two constituents of RGL types. It is a deconstructed $\mathsf{Cl}$ where the subject NP is separated from the rest of the clause. The motivation for this is to allow for nested frames (see Section~\ref{ssec:phrasebook}) and for adding non-core FEs before combining the NP and VP parts into a clause (see Section~\ref{ssec:museum}).

The RGL-subcategorized FEs of the shared valence patterns are declared as common semantic types (categories). Although the conception of BFN states that core FEs are unique to the frame, even though their names are not unique across frames, we do not make such a distinction at the level of types; they are implicitly made frame-specific by the frame functions. The only distinction is based on the syntactic realization.

In order to keep the FE names unique, the RGL types are added as suffixes:

\begin{enumerate}
\item[] $\mathit{cat}$ $\mathsf{Event\_VP}$
\item[] $\mathit{cat}$ $\mathsf{Experiencer\_NP}$
\item[] $\mathit{cat}$ $\mathsf{Focal\_participant\_Adv}$
\item[] $\mathit{cat}$ $\mathsf{Focal\_participant\_NP}$
\end{enumerate}

Note that the {\sc Focal\_participant} is typically realized as a noun phrase, but some intransitive \texttt{Desiring} verbs require it as a prepositional phrase (PP), hence this FE is subcategorized using the RGL types $\mathsf{NP}$ and $\mathsf{Adv}$ (adverbial modifier). In GF, the type $\mathsf{Adv}$ covers both adverbs and PPs, and there is no separate type for PPs. Also note that all FEs are specified as optional arguments in the concrete syntaxes, i.e. any FE can be an empty phrase if it is not expressed in the sentence.

The frame-evoking target verb is always expected as the last, mandatory argument. We assume that verbs of the same type evoking the same frame share, in general, a subset of normalized semantico-syntactic valence patterns of that frame. Patterns requiring, for instance, a transitive verb cannot be evoked by an intransitive verb. Otherwise, the current approach does not limit the set of verbs that can evoke a frame, and the set of prepositions that can be used for an FE if it is realized as a PP. We expect that appropriate verbs and prepositions are specified by the application grammar that uses the FrameNet-based grammar as an API. Hence, this approach allows evoking a frame by a metaphor, i.e. an LU that normally evokes another frame.

The design and implementation of the abstract and concrete syntaxes of lexical entries is described in Section~\ref{sec:lexicon}.

\subsection{Concrete Syntaxes}
\label{ssec:concrete}

The exact behaviour (linearization) of types and functions declared in the abstract syntax is defined in the concrete syntax of each language.

The mapping from the semantic BFN types (FEs) to the syntactic RGL types is straightforward and is shared for all languages in a functor, for instance:

\begin{enumerate}
\item[] $\mathit{lincat}$ $\mathsf{Event\_VP = Maybe}$ $\mathsf{VP}$
\item[] $\mathit{lincat}$ $\mathsf{Focal\_participant\_NP = Maybe}$ $\mathsf{NP}$
\item[] $\mathit{lincat}$ $\mathsf{Focal\_participant\_Adv = Maybe}$ $\mathsf{Adv}$
\end{enumerate}

To allow for optional FEs (verb arguments that might not be expressed in the sentence), all linearization types are of type $\mathsf{Maybe}$ whose behaviour is similar to the analogous type in Haskell: a value of type $\mathsf{Maybe}$ $\mathit{x}$ either contains a value of type $\mathit{x}$ (represented as $\mathsf{Just}$ $\mathit{x}$), or it is not provided (represented as $\mathsf{Nothing}$).

To implement the frame functions, particularly, to fill the verb phrase part of $\mathsf{Clause}$ objects, RGL constructors are applied to the arguments depending on their grammatical types and relations, and the grammatical voice. The implementation of functions declared in Section~\ref{sssec:implement} is systematically generated for English and Swedish as follows:

\begin{enumerate}
\item[] $\mathit{lin}$ $\mathsf{Desiring\_V}$ $\mathrm{experiencer\_np}$ $\mathrm{focal\_participant\_adv}$ $\mathrm{v = \{}$
\begin{enumerate}
\item[] $\mathrm{np =}$ $\mathsf{fromMaybe}$ $\mathsf{NP}$ $\mathsf{emptyNP}$ $\mathrm{experiencer\_np}$ $\mathrm{;}$
\item[] $\mathrm{vp =}$ $\mathsf{mkVP}$ $\mathsf{(mkVP}$ $\mathrm{v}$$\mathsf{)}$ $\mathsf{(fromMaybe}$ $\mathsf{Adv}$ $\mathsf{emptyAdv}$ $\mathrm{focal\_participant\_adv}$$\mathsf{)}$ $\mathrm{\}}$
\end{enumerate}
\end{enumerate}

\begin{enumerate}
\item[] $\mathit{lin}$ $\mathsf{Desiring\_VV}$ $\mathrm{event\_vp}$ $\mathrm{experiencer\_np}$ $\mathrm{vv = \{}$
\begin{enumerate}
\item[] $\mathrm{np =}$ $\mathsf{fromMaybe}$ $\mathsf{NP}$ $\mathsf{emptyNP}$ $\mathrm{experiencer\_np}$ $\mathrm{;}$
\item[] $\mathrm{vp =}$ $\mathsf{mkVP}$ $\mathrm{vv}$ $\mathsf{(fromMaybe}$ $\mathsf{VP}$ $\mathsf{emptyVP}$ $\mathrm{event\_vp}$$\mathsf{)}$ $\mathrm{\}}$
\end{enumerate}
\end{enumerate}

\begin{enumerate}
\item[] $\mathit{lin}$ $\mathsf{Desiring\_V2}$ $\mathrm{experiencer\_np}$ $\mathrm{focal\_participant\_np}$ $\mathrm{v2 = \{}$
\begin{enumerate}
\item[] $\mathrm{np =}$ $\mathsf{fromMaybe}$ $\mathsf{NP}$ $\mathsf{emptyNP}$ $\mathsf{experiencer\_np}$ $\mathrm{;}$
\item[] $\mathrm{vp =}$ $\mathsf{mkVP}$ $\mathrm{v2}$ $\mathsf{(fromMaybe}$ $\mathsf{NP}$ $\mathsf{emptyNP}$ $\mathrm{focal\_participant\_np}$$\mathsf{)}$ $\mathrm{\}}$
\end{enumerate}
\end{enumerate}

\begin{enumerate}
\item[] $\mathit{lin}$ $\mathsf{Desiring\_V2\_Pass}$ $\mathrm{experiencer\_np}$ $\mathrm{focal\_participant\_np}$ $\mathrm{v2 = \{}$
\begin{enumerate}
\item[] $\mathrm{np =}$ $\mathsf{fromMaybe}$ $\mathsf{NP}$ $\mathsf{emptyNP}$ $\mathrm{focal\_participant\_np}$ $\mathrm{;}$
\item[] $\mathrm{vp =}$ $\mathsf{mkVP}$
\begin{enumerate}
\item[] $\mathsf{(passiveVP}$ $\mathrm{v2}$$\mathsf{)}$
\item[] $\mathsf{(mkAdv}$ $\mathsf{by8agent\_Prep}$ $\mathsf{(fromMaybe}$ $\mathsf{NP}$ $\mathsf{emptyNP}$ $\mathrm{experiencer\_np}$$\mathsf{))}$ $\mathrm{\}}$
\end{enumerate}
\end{enumerate}
\end{enumerate}

To the $\mathsf{NP}$ field of a $\mathsf{Clause}$ object, either the value of the corresponding $\mathsf{NP_{nsubj}}$ or $\mathsf{NP_{nsubjpass}}$ argument, or an empty string of type $\mathsf{NP}$ is assigned. This choice is handled by the helper function $\mathsf{Maybe.fromMaybe}$ that takes a $\mathsf{Maybe}$ value and returns a predefined empty phrase of the respective type if the $\mathsf{Maybe}$ value is not provided ($\mathsf{Nothing}$); otherwise it returns the provided value. Optional verb complements are handled similarly.

In order to produce a value of the $\mathsf{VP}$ field, RGL constructors $\mathsf{mkVP}$, $\mathsf{passiveVP}$, $\mathsf{mkAdv}$ etc. and RGL structural words $\mathsf{by8agent\_Prep}$ (prepositions \emph{by} and \emph{av} in English and Swedish respectively) etc. are applied, for instance:\footnote{\url{http://www.grammaticalframework.org/lib/doc/synopsis.html}}

\begin{enumerate}
\item[] $\mathsf{mkVP : V \to VP}$
\item[] $\mathsf{mkVP : V2 \to NP \to VP}$
\item[] $\mathsf{mkVP : VV \to VP \to VP}$
\item[] $\mathsf{mkVP : VP \to Adv \to VP}$
\end{enumerate}

The RGL-based code templates used to implement the above functions can be systematically reused for many other frame functions. Given the set of shared valence patterns, there are only 32 syntactic patterns that cover all 869 semantico-syntactic patterns (Table~\ref{tab:freq}). By syntactic valence patterns we mean patterns that specify only the grammatical types and relations of FEs, and the grammatical voice. As Table~\ref{tab:freq} shows, the syntactic patterns underlying functions $\mathsf{Desiring\_V2}$, $\mathsf{Desiring\_V}$, $\mathsf{Desiring\_VV}$ and $\mathsf{Desiring\_V2\_Pass}$ already cover more that 54\% of all the shared frame functions. For the same verb types ($\mathsf{V}$, $\mathsf{V2}$, $\mathsf{VV}$), other syntactic patterns cover another 39\% of frame functions for which the code templates are derived in several ways:

\begin{itemize}[noitemsep]
\item complements of $\mathsf{Adv}$ type are added by recursively applying the respective $\mathsf{mkVP}$ constructor, or they are eliminated at all;\footnote{The order of $\mathsf{Adv}$ complements is based on the most frequent sentence pattern.}
\item the $\mathsf{NP}$ field of $\mathsf{Clause}$ is fixed to the empty string if the valence pattern does not include the subject FE (e.g. due to examples only in the imperative mood);\footnote{A missing subject FE, however, could be often automatically inferred and added.}
\item the agent FE that would be the subject in the active voice but is missing in the passive voice is fixed to the empty string.
\end{itemize}

\begin{table}[h]
\tabcolsep 4pt
\begin{center}
\begin{tabular}{|lllr||lllr|}
\hline
Verb & Voice & Arguments & Freq. & Verb & Voice & Arguments & Freq. \\
\hline
$\mathsf{V2}$ & $\mathsf{Act}$ & $\mathsf{NP_{dobj}}$ $\mathsf{NP_{nsubj}}$ & 277 & $\mathsf{V}$ & $\mathsf{Act}$ & $\mathsf{Adv}$ $\mathsf{Adv}$ $\mathsf{Adv}$ $\mathsf{NP_{nsubj}}$ & 2\\
$\mathsf{V}$ & $\mathsf{Act}$ & $\mathsf{Adv}$ $\mathsf{NP_{nsubj}}$ & 155 & $\mathsf{V2}$ & $\mathsf{Act}$ & $\mathsf{Adv}$ $\mathsf{Adv}$ $\mathsf{NP_{dobj}}$ $\mathsf{NP_{nsubj}}$ & 2\\
$\mathsf{V2}$ & $\mathsf{Pass}$ & $\mathsf{NP_{nsubjpass}}$ & 84 & $\mathsf{V3}$ & $\mathsf{Act}$ & $\mathsf{NP_{iobj}}$ $\mathsf{NP_{nsubj}}$ & 2\\
$\mathsf{V2}$ & $\mathsf{Act}$ & $\mathsf{Adv}$ $\mathsf{NP_{dobj}}$ $\mathsf{NP_{nsubj}}$ & 80 & $\mathsf{VQ}$ & $\mathsf{Act}$ & $\mathsf{QS}$ & 2\\
$\mathsf{V}$ & $\mathsf{Act}$ & $\mathsf{NP_{nsubj}}$ & 78 & $\mathsf{VS}$ & $\mathsf{Act}$ & $\mathsf{Adv}$ $\mathsf{NP_{nsubj}}$ $\mathsf{S}$ & 2\\
$\mathsf{V2}$ & $\mathsf{Pass}$ & $\mathsf{Adv}$ $\mathsf{NP_{nsubjpass}}$ & 34 & $\mathsf{V2}$ & $\mathsf{Pass}$ & $\mathsf{Adv}$ $\mathsf{Adv}$ $\mathsf{NP_{nsubjpass}}$ & 2\\
$\mathsf{VS}$ & $\mathsf{Act}$ & $\mathsf{NP_{nsubj}}$ $\mathsf{S}$ & 29 & $\mathsf{V2}$ & $\mathsf{Pass}$ & $\mathsf{Adv}$ $\mathsf{NP_{dobj}}$ $\mathsf{NP_{nsubjpass}}$ & 2\\
$\mathsf{VV}$ & $\mathsf{Act}$ & $\mathsf{NP_{nsubj}}$ $\mathsf{VP}$ & 21 & $\mathsf{V2}$ & $\mathsf{Pass}$ & $\mathsf{NP_{dobj}}$ & 2\\
$\mathsf{V2}$ & $\mathsf{Pass}$ & $\mathsf{NP_{dobj}}$ $\mathsf{NP_{nsubjpass}}$ & 19 & $\mathsf{V2}$ & $\mathsf{Act}$ & $\mathsf{Adv}$ $\mathsf{Adv}$ $\mathsf{NP_{dobj}}$ & 1\\
$\mathsf{V2}$ & $\mathsf{Act}$ & $\mathsf{NP_{dobj}}$ & 17 & $\mathsf{V2S}$ & $\mathsf{Act}$ & $\mathsf{NP_{dobj}}$ $\mathsf{NP_{nsubj}}$ $\mathsf{S}$ & 1\\
$\mathsf{V}$ & $\mathsf{Act}$ & $\mathsf{Adv}$ $\mathsf{Adv}$ $\mathsf{NP_{nsubj}}$ & 16 & $\mathsf{V2S}$ & $\mathsf{Act}$ & $\mathsf{NP_{dobj}}$ $\mathsf{S}$ & 1\\
$\mathsf{VQ}$ & $\mathsf{Act}$ & $\mathsf{NP_{nsubj}}$ $\mathsf{QS}$ & 10 & $\mathsf{V2V}$ & $\mathsf{Act}$ & $\mathsf{NP_{dobj}}$ $\mathsf{VP}$ & 1\\
$\mathsf{V2}$ & $\mathsf{Act}$ & $\mathsf{Adv}$ $\mathsf{NP_{dobj}}$ & 9 & $\mathsf{VS}$ & $\mathsf{Act}$ & $\mathsf{S}$ & 1\\
$\mathsf{V}$ & $\mathsf{Act}$ & $\mathsf{Adv}$ & 8 & $\mathsf{VV}$ & $\mathsf{Act}$ & $\mathsf{VP}$ & 1\\
$\mathsf{V2V}$ & $\mathsf{Act}$ & $\mathsf{NP_{dobj}}$ $\mathsf{NP_{nsubj}}$ $\mathsf{VP}$ & 5 & $\mathsf{V2}$ & $\mathsf{Pass}$ & $\mathsf{Adv}$ & 1\\
$\mathsf{VS}$ & $\mathsf{Pass}$ & $\mathsf{S}$ & 3 & $\mathsf{VS}$ & $\mathsf{Pass}$ & $\mathsf{NP_{nsubjpass}}$ $\mathsf{S}$ & 1\\
\hline
\end{tabular}
\end{center}
\caption{\label{tab:freq} Syntactic valence patterns underlying the shared semantico-syntactic patterns. The order of arguments (FEs) is not taken into account.}
\end{table}

The remaining less than 7\% of the shared frame functions represent the use of other verb types~-- $\mathsf{VS}$, $\mathsf{VQ}$, $\mathsf{V2V}$, $\mathsf{V3}$ and $\mathsf{V2S}$~-- for which the respective RGL constructors are applied:

\begin{enumerate}
\item[] $\mathsf{mkVP : VS \to S \to VP}$
\item[] [\emph{I}]\textsubscript{$\mathsf{Cognizer/NP}$} \emph{do} [\emph{REMEMBER}]\textsubscript{$\mathsf{VS}$} [\emph{we did a few gigs}]\textsubscript{$\mathsf{Content/S}$}
\end{enumerate}
\begin{enumerate}
\item[] $\mathsf{mkVP : VQ \to QS \to VP}$
\item[] [\emph{he}]\textsubscript{$\mathsf{Cognizer/NP}$} [\emph{RECOGNIZED}]\textsubscript{$\mathsf{VS}$} [\emph{where he was}]\textsubscript{$\mathsf{Phenomenon/QS}$}
\end{enumerate}
\begin{enumerate}
\item[] $\mathsf{mkVP : V2V \to NP \to VP \to VP}$
\item[] [\emph{you}]\textsubscript{$\mathsf{Speaker/NP}$} \emph{specifically} [\emph{REQUEST}]\textsubscript{$\mathsf{V2V}$} [\emph{me}]\textsubscript{$\mathsf{Addressee/NP}$} [\emph{to do so}]\textsubscript{$\mathsf{Message/VP}$}
\end{enumerate}
\begin{enumerate}
\item[] $\mathsf{mkVP : V3 \to NP \to NP \to VP}$
\item[] [\emph{you}]\textsubscript{$\mathsf{Agent/NP}$} [\emph{DENIED}]\textsubscript{$\mathsf{V3}$} [\emph{her}]\textsubscript{$\mathsf{Protagonist/NP}$} [\emph{any life of her own}]\textsubscript{$\mathsf{State\_of\_affairs/NP}$}
\end{enumerate}
\begin{enumerate}
\item[] $\mathsf{mkVP : V2S \to NP \to S \to VP}$
\item[] [\emph{he}]\textsubscript{$\mathsf{Speaker/NP}$} [\emph{PERSUADED}]\textsubscript{$\mathsf{V2S}$} [\emph{himself}]\textsubscript{$\mathsf{Addressee/NP}$} [\emph{that they helped}]\textsubscript{$\mathsf{Content/S}$}
\end{enumerate}

Note that the type $\mathsf{S}$, an embedded declarative sentence, is used only if the subclause can be paraphrased using the subjunction ($\mathsf{Subj}$) \emph{that}; otherwise such FEs are subcategorized as $\mathsf{Adv}$, and the application grammar has to specify the subjunction by applying the RGL constructor $\mathsf{mkAdv : Subj \to S \to Adv}$.

Also note that FEs of type $\mathsf{VP}$, $\mathsf{S}$ and $\mathsf{QS}$, and $\mathsf{Adv}$ encapsulating $\mathsf{S}$ represent nested frames. We use the types $\mathsf{S}$ and $\mathsf{QS}$ instead of $\mathsf{Cl}$ and $\mathsf{QCl}$ to allow for specifying sentence level parameters like tense, anteriority and polarity of the nested frames.

The implementation of frame functions, although currently kept separate for each language, mostly could be shared in a functor thanks to the syntactic abstraction provided by RGL. In general, however, the order of $\mathsf{Adv}$ FEs differ across languages.

\section{FrameNet-Based Lexicon}
\label{sec:lexicon}

In GF, there is no formal distinction between syntactic rules and lexical entries. Lexical entries are represented by functions that normally take no arguments and usually but not necessarily return values of lexical categories (e.g. $\mathsf{V}$ versus $\mathsf{VP}$).

LUs between BFN and SweFN (and other framenets) are not explicitly aligned, therefore we first extract and generate a framenet-specific lexicon for each language. Second, we have conducted an experiment to automatically produce a shared lexicon by partially aligning LUs between BFN and SweFN.

\subsection{Abstract Syntaxes}
\label{ssec:lex-abstract}

Following the design of the FrameNet-based grammar, LUs in our approach are subcategorized by GF RGL verb types, therefore for each LU there is one or more lexical entry in the lexicon.

The abstract lexical identifiers (function names) start with the language-specific base form of the verb. To distinguish between different types and senses of LUs, the verb type and the frame name is appended to the identifiers as illustrated in Tables~\ref{tab:eng} and \ref{tab:swe}.

\begin{table}[b]
\tabcolsep 3.75pt
\begin{center}
\begin{tabular}{|l|c|l|}
\hline
Function & Type & Annotated corpus excerpt\\
\hline
$\mathsf{feel\_like\_V2\_Desiring}$ & $\mathsf{V2}$ & [\emph{I}]\textsubscript{$\mathsf{Experiencer/NP}$} [\emph{FEEL LIKE}]\textsubscript{$\mathsf{V2}$} [\emph{a glass}]\textsubscript{$\mathsf{Focal\_participant/NP}$}\\
$\mathsf{feel\_like\_VV\_Desiring}$ & $\mathsf{VV}$ & [\emph{I}]\textsubscript{$\mathsf{Experiencer/NP}$} [\emph{FELT LIKE}]\textsubscript{$\mathsf{VV}$} [\emph{shouting}]\textsubscript{$\mathsf{Event/VP}$}\\
$\mathsf{want\_V\_Desiring}$ & $\mathsf{V}$ & [\emph{he}]\textsubscript{$\mathsf{Experiencer/NP}$} [\emph{WANTED}]\textsubscript{$\mathsf{V}$} [\emph{more}]\textsubscript{$\mathsf{Focal\_participant/Adv}$}\\
$\mathsf{want\_V2\_Desiring}$ & $\mathsf{V2}$ & [\emph{you}]\textsubscript{$\mathsf{Experiencer/NP}$} [\emph{WANT}]\textsubscript{$\mathsf{V2}$} [\emph{one}]\textsubscript{$\mathsf{Focal\_participant/NP}$}\\
$\mathsf{want\_VV\_Desiring}$ & $\mathsf{VV}$ & [\emph{I}]\textsubscript{$\mathsf{Experiencer/NP}$} \emph{would n't} [\emph{WANT}]\textsubscript{$\mathsf{VV}$} [\emph{to know}]\textsubscript{$\mathsf{Event/VP}$}\\
$\mathsf{yearn\_V\_Desiring}$ & $\mathsf{V}$ & [\emph{he}]\textsubscript{$\mathsf{Experiencer/NP}$} \emph{'d} [\emph{YEARN}]\textsubscript{$\mathsf{V}$} [\emph{for England}]\textsubscript{$\mathsf{Focal\_participant/Adv}$}\\
$\mathsf{yearn\_VV\_Desiring}$ & $\mathsf{VV}$ & [\emph{he}]\textsubscript{$\mathsf{Experiencer/NP}$} \emph{'d} [\emph{YEARNED}]\textsubscript{$\mathsf{VV}$} [\emph{to phone Liz}]\textsubscript{$\mathsf{Event/VP}$}\\
\hline
\end{tabular}
\end{center}
\caption{\label{tab:eng} Sample lexical entries extracted from the BFN corpus.}
\end{table}

\begin{table}[b]
\tabcolsep 1pt
\begin{center}
\begin{tabular}{|l|c|l|}
\hline
Function & Type & Annotated corpus excerpt\\
\hline
$\mathsf{k\ddot{a}nna\_V2\_Awareness}$ & $\mathsf{V2}$ & [\emph{vi}]\textsubscript{$\mathsf{Cognizer/NP}$} \emph{inte} [\emph{KÄNNER}]\textsubscript{$\mathsf{V2}$} [\emph{orsaken till}]\textsubscript{$\mathsf{Content/NP}$}\\ 
$\mathsf{k\ddot{a}nna\_V2\_Familiarity}$ & $\mathsf{V2}$ & [\emph{jag}]\textsubscript{$\mathsf{Cognizer/NP}$} [\emph{KÄNNER}]\textsubscript{$\mathsf{V2}$} [\emph{Eva}]\textsubscript{$\mathsf{Entity/NP}$}\\
$\mathsf{k\ddot{a}nna\_f\ddot{o}r\_V2\_Desiring}$ & $\mathsf{V2}$ & [\emph{jag}]\textsubscript{$\mathsf{Experiencer/NP}$} [\emph{KÄNNER FÖR}]\textsubscript{$\mathsf{V2}$} [\emph{en tur}]\textsubscript{$\mathsf{Focal\_participant/NP}$}\\ 
$\mathsf{k\ddot{a}nna\_f\ddot{o}r\_VV\_Desiring}$ & $\mathsf{VV}$ & [\emph{jag}]\textsubscript{$\mathsf{Experiencer/NP}$} [\emph{KÄNNER FÖR}]\textsubscript{$\mathsf{VV}$} [\emph{att skriva en bok}]\textsubscript{$\mathsf{Event/VP}$}\\ 
$\mathsf{k\ddot{a}nna\_sig\_V\_Feeling}$ & $\mathsf{V}$ & [\emph{man}]\textsubscript{$\mathsf{Experiencer/NP}$} [\emph{KÄNNER SIG}]\textsubscript{$\mathsf{V}$} [\emph{trygg}]\textsubscript{$\mathsf{Emotional\_state/Adv}$}\\
$\mathsf{l\ddot{a}ngta\_V\_Desiring}$ & $\mathsf{V}$ & [\emph{Roberte}]\textsubscript{$\mathsf{Experiencer/NP}$} [\emph{LÄNGTADE}]\textsubscript{$\mathsf{V}$} [\emph{hem}]\textsubscript{$\mathsf{Focal\_participant/Adv}$}\\ 
$\mathsf{vilja\_VV\_Desiring}$ & $\mathsf{VV}$ &
[\emph{jag}]\textsubscript{$\mathsf{Experiencer/NP}$} [\emph{VILJA}]\textsubscript{$\mathsf{VV}$} [\emph{ha sju sångare}]\textsubscript{$\mathsf{Event/VP}$}\\ 
\hline
\end{tabular}
\end{center}
\caption{\label{tab:swe} Sample lexical entries extracted from the SweFN corpus.}
\end{table}

The generation of the abstract language-specific lexicons is straightforward. Given the set of 869 shared valence patterns (Section~\ref{sssec:shared}), we select all the distinct target verbs from the sentence patterns (Section~\ref{sssec:patterns}) that belong to the shared patterns. Then we append the corresponding verb type and frame name to the base form of the target verb and declare all the resulting identifiers as nullary functions returning verbs of the respective types.

From the BFN corpus, we have extracted 2,831 LUs resulting in 3,432 lexical entries (due to alternative verb types). For Swedish, the numbers are 1,844 and 1,899 respectively. The ratio of lexical entries per LU is considerably smaller for Swedish (1.03 versus 1.21) because of the small number of SweFN examples per LU (around 1.5 versus around 20 in BFN; see Tables~\ref{bfn_statistics} and \ref{swefn_statistics} in Section~\ref{sssec:series}).

\subsection{Concrete Syntaxes}
\label{ssec:lex-concrete}

In order to generate concrete lexicons, first, we have to specify an appropriate inflectional paradigm for each verb independently of its potential senses (frames) and valence types. Inflectional paradigms are represented by language-specific constructors provided in the RGL \texttt{Paradigms\emph{L}} modules. Each constructor, which can be overloaded, expects specific verb forms as arguments from which all forms of the verb can be generated, for instance:

\begin{enumerate}
\item[] $\mathsf{irregV}$ ``\emph{feel}'' ``\emph{felt}'' ``\emph{felt}''
\item[] $\mathsf{regV}$ ``\emph{want}''
\item[] $\mathsf{mkV}$ ``\emph{yearn}'' ``\emph{yearns}'' ``\emph{yearned}'' ``\emph{yearned}'' ``\emph{yearning}''
\end{enumerate}

\begin{enumerate}
\item[] $\mathsf{irregV}$ ``känna'' ``kände'' ``känt''
\item[] $\mathsf{mkV}$ ``längtar''
\item[] $\mathsf{mkV}$ ``vilja'' ``vill'' ``vilj'' ``ville'' ``velat'' ``velad''
\end{enumerate}

The first argument usually is the base form, but it can be another form from which the base form can be straightforwardly derived (e.g. \emph{längtar} `[one] longs').

We extract such verb-constructor pairs from the existing monolingual and multilingual RGL dictionaries and other modules (in the reverse order of preference):

\begin{enumerate}[leftmargin=1cm]
\item \texttt{\emph{L}/Dict\emph{L}} (6,034 pairs for English, 7,324 for Swedish)
\item \texttt{translator/Dictionary\emph{L}} (6,037 pairs for English, 2,430 for Swedish)
\item \texttt{\emph{L}/Lexicon\emph{L}} (98 pairs for English, 96 for Swedish)
\item \texttt{\emph{L}/Irreg\emph{L}} (173 pairs for English, 182 for Swedish)
\item \texttt{\emph{L}/Structural\emph{L}} (2 pairs for English, 4 for Swedish)
\end{enumerate}

In total, we have extracted constructors for 6,040 English verbs and 7,492 Swedish verbs. Still, 59 BFN verbs and 28 SweFN verbs are out-of-vocabulary.\footnote{The RGL modules \texttt{Dict\emph{L}}, \texttt{Dictionary\emph{L}}, \texttt{Lexicon\emph{L}}, \texttt{Irreg\emph{L}} and \texttt{Structural\emph{L}} are a subject to change independently. We have used an RGL snapshot of December 2014.}

Second, for each lexical entry, we generate its linearization rule based on (i) the corresponding verb constructor, (ii) particles and reflexive pronouns, if any, that constitute the LU and (iii) the verb type of the lexical entry, for instance:

\begingroup
\fontsize{9.25pt}{11.25pt}\selectfont
\begin{enumerate}
\item[] $\mathit{lin}$ $\mathsf{want\_V2\_Desiring = mkV2}$ $\mathsf{(regV}$ ``\emph{want}''$\mathsf{)}$
\end{enumerate}
\endgroup

\begingroup
\fontsize{9.2pt}{11.2pt}\selectfont
\begin{enumerate}
\item[] $\mathit{lin}$ $\mathsf{k\ddot{a}nna\_f\ddot{o}r\_VV\_Desiring = mkVV}$ $\mathsf{(partV}$ $\mathsf{(irregV}$ ``\emph{känna}'' ``\emph{kände}'' ``\emph{känt}''$\mathsf{)}$ ``\emph{för}''$\mathsf{)}$
\item[] $\mathit{lin}$ $\mathsf{k\ddot{a}nna\_sig\_V\_Feeling = reflV}$ $\mathsf{(irregV}$ ``\emph{känna}'' ``\emph{kände}'' ``\emph{känt}''$\mathsf{)}$
\end{enumerate}
\endgroup

In the result, we were able to generate linearization rules for currently 3,350 (98\%) out of 3,432 BFN entries and for 1,789 (94\%) out of 1,899 SweFN entries.

At this point, it should be noted that each sentence pattern (Section~\ref{sssec:patterns}) includes not only a reference to the LU but also a morphological description of the LU constituents, which is important in the case of multi-word expressions (MWE), e.g. \emph{feel like}, \emph{känna för}, \emph{känna sig} etc. Moreover, the morphological descriptions are unified across languages according to the universal POS tags\footnote{\url{http://universaldependencies.github.io/docs/u/pos/}} and features\footnote{\url{http://universaldependencies.github.io/docs/u/feat/}} allowing for a common generator of concrete lexicons.

The current approach to the FrameNet-based grammar and lexicon supports linearization of relatively simple MWEs that, apart from the main verb, include particles (constructor $\mathsf{partV}$) and reflexive pronouns (constructor $\mathsf{reflV}$) in any combination.

Considering only the shared frame valence patterns, we have extracted 98 such lexical entries for English, which is about 3\% of all entries extracted from BFN and about 84\% of all MWE entries extracted from BFN. All these entries correspond to the same morphological pattern:
\begin{enumerate}
\item[] $\mathsf{VERB.Fin}$~~$\mathsf{ADP}$
\end{enumerate}

\noindent where $\mathsf{ADP}$ (adposition) represents a particle. For Swedish, we have extracted 465 such entries, which is about 25\% of all entries and about 85\% of all MWE entries extracted from SweFN. In addition to the MWE pattern found in BFN, SweFN covers several other patterns of simple MWEs:

\begin{enumerate}
\item[] $\mathsf{VERB.Fin}$~~$\mathsf{ADP}$~~$\mathsf{ADP}$
\item[] $\mathsf{VERB.Fin}$~~$\mathsf{ADP}$~~$\mathsf{PRON.Reflex}$
\item[] $\mathsf{VERB.Fin}$~~$\mathsf{PRON.Reflex}$
\item[] $\mathsf{VERB.Fin}$~~$\mathsf{PRON.Reflex}$~~$\mathsf{ADP}$
\end{enumerate}

Patterns of currently unsupported, more complex MWEs are summarized in Table~\ref{tab:complex-mwe}. This leads to 19 MWE entries in the English lexicon and 82 MWE entries in the Swedish lexicon having no linearization.

\begin{table}[b]
\tabcolsep 4pt
\begin{center}
\begin{tabular}{|l|c|c|}
\hline
Patterns of complex MWEs & BFN & SweFN\\
\hline
$\mathsf{VERB.Fin}$~~$\mathsf{ADJ}$ & \texttt{x} & \texttt{x}\\
$\mathsf{VERB.Fin}$~~$\mathsf{ADP}$~~$\mathsf{ADP}$~~$\mathsf{NOUN}$ & ~ & \texttt{x}\\
$\mathsf{VERB.Fin}$~~$\mathsf{ADP}$~~$\mathsf{NOUN}$ & \texttt{x} & \texttt{x}\\
$\mathsf{VERB.Fin}$~~$\mathsf{ADP}$~~$\mathsf{PRON.Prs}$ & ~ & \texttt{x}\\
$\mathsf{VERB.Fin}$~~$\mathsf{ADP}$~~$\mathsf{VERB.Fin}$ & ~ & \texttt{x}\\
$\mathsf{VERB.Fin}$~~$\mathsf{DET.Art.Def}$~~$\mathsf{NOUN}$ & \texttt{x} & ~\\
$\mathsf{VERB.Fin}$~~$\mathsf{DET.Art.Ind}$~~$\mathsf{NOUN}$ & ~ & \texttt{x}\\
$\mathsf{VERB.Fin}$~~$\mathsf{NOUN}$ & \texttt{x} & \texttt{x}\\
$\mathsf{VERB.Fin}$~~$\mathsf{NOUN}$~~$\mathsf{ADP}$ & \texttt{x} & \texttt{x}\\
$\mathsf{VERB.Fin}$~~$\mathsf{NOUN}$~~$\mathsf{DET.Art.Def}$~~$\mathsf{NOUN}$ & \texttt{x} & ~\\
$\mathsf{VERB.Fin}$~~$\mathsf{PRON.Reflex}$~~$\mathsf{ADJ}$ & ~ & \texttt{x}\\
$\mathsf{VERB.Fin}$~~$\mathsf{PRON.Reflex}$~~$\mathsf{ADP}$~~$\mathsf{NOUN}$ & ~ & \texttt{x}\\
$\mathsf{VERB.Fin}$~~$\mathsf{VERB.Fin}$ & \texttt{x} & \texttt{x}\\
\hline
\end{tabular}
\end{center}
\caption{\label{tab:complex-mwe} Morphological patterns of MWE entries that, in the current implementation of the FrameNet-based grammar and lexicon, have no linearization rules.}
\end{table}

To address this issue, we could include lexical entries of type $\mathsf{VP}$ implying a similar syntactic valence as for verbs of type $\mathsf{V}$. However, this would require to introduce separate frame functions. An alternative approach would be to extend the notion and support of particle verbs in RGL so that ``particles'' could be involved in the syntactic agreement.

\subsection{Aligning Lexical Units Across Languages}
\label{ssec:lex-mapping}

The multilingual RGL lexicons~-- the large translation dictionary (modules \texttt{Dictio}-\texttt{nary\emph{L}}) and the small lexicon of frequently used words (modules \texttt{Lexicon\emph{L}})~-- can be used not only for the extraction of verb constructors but also for aligning LUs (i.e. lexical entries) across languages.

Let us consider the following example. For the frame \texttt{Desiring}, we have extracted several lexical entries of type $\mathsf{VV}$ as shown in Table~\ref{tab:eng} for English and in Table~\ref{tab:swe} for Swedish.

If we search for the English verbs \emph{feel}, \emph{want} and \emph{yearn}, and for the Swedish verbs \emph{känna} and \emph{vilja} in the RGL modules \texttt{DictionaryEng} and \texttt{DictionarySwe} respectively, we find these mappings (among others):

\begin{enumerate}
\item[] \texttt{DictionaryEng}: $\mathit{lin}$ $\mathsf{feel\_V = IrregEng.feel\_V}$
\item[] \texttt{DictionarySwe}: $\mathit{lin}$ $\mathsf{feel\_V = mkV}$ ``känna'' ``kände'' ``känt''
\end{enumerate}

\begin{enumerate}
\item[] \texttt{DictionaryEng}: $\mathit{lin}$ $\mathsf{want\_V2 = mkV2}$ $\mathsf{(mkV}$ ``want''$\mathsf{)}$
\item[] \texttt{DictionarySwe}: $\mathit{lin}$ $\mathsf{want\_V2 = mkV2}$ $\mathsf{IrregSwe.vilja\_V}$
\end{enumerate}

\begin{enumerate}
\item[] \texttt{DictionaryEng}: $\mathit{lin}$ $\mathsf{yearn\_V = mkV}$ ``yearn'' ``yearns'' ``yearned'' ...
\item[] \texttt{DictionarySwe}: $\mathit{lin}$ $\mathsf{yearn\_V = mkV}$ ``trängtar''
\end{enumerate}

\noindent suggesting the following alignment between the framenet-specific lexicons:

\begin{enumerate}
\item[] $\mathsf{feel\_like\_VV\_Desiring = k\ddot{a}nna\_f\ddot{o}r\_VV\_Desiring}$
\item[] $\mathsf{want\_VV\_Desiring = vilja\_VV\_Desiring}$
\end{enumerate}

We have collected all such suggestions in a separate shared lexicon where BNF identifiers are used as interlingua symbols in the abstract syntax, and the framenet-specific lexicons are used as resource libraries to implement the concrete syntaxes.

The generation of the concrete English lexicon is trivial, for instance:
\begin{enumerate}
\item[] $\mathit{lin}$ $\mathsf{want\_VV\_Desiring = want\_VV\_Desiring}$
\end{enumerate}

The concrete Swedish lexicon is generated as illustrated in the alignment example above, and it can include alternative variants, for instance:
\begin{enumerate}
\item[] $\mathit{lin}$ $\mathsf{know\_V2\_Familiarity = }$
\begin{enumerate}
\item[] $\mathsf{variants}$ $\mathsf{\{k\ddot{a}nna\_V2\_Familiarity}$ $\mathsf{|}$ $\mathsf{k\ddot{a}nna\_till\_V2\_Familiarity\}}$
\end{enumerate}
\end{enumerate}

\noindent meaning that all variants will be considered while parsing a sentence, but only the first variant will be used for linearization. Currently, variants are ordered so that MWEs follow simple verbs, otherwise they are given in the alphabetical order; however, they should be ordered at least by frequency.

In the case of MWEs, we search for alignment variants based on the main verb if there is no match for the whole MWE. This improves the coverage (as it is illustrated with \emph{feel like} above) but sometimes leads to incorrect alignments, for instance, \emph{exhale} has been aligned with \emph{andas in} `inhale':
\begin{enumerate}
\item[] $\mathit{lin}$ $\mathsf{exhale\_V2\_Breathing = andas\_in\_V2\_Breathing}$
\end{enumerate}

In the result, we have aligned 703 BFN entries (21\%) with 900 SweFN entires (47\%). This approach is still promising, and there is a clear space for improvement:

\begin{enumerate}[leftmargin=1cm]
\item The alignment procedure failed for about 30\% of BFN entries because of missing linearization for nearly 800 \texttt{DictionarySwe} entries.
\item For nearly half of BFN entries, alignment was not found because no match was found among SwFN entries of the same type belonging to the same frame, which is a consequence of the comparatively small size of SweFN (2.2 SweFN entries versus 4 BFN entries per shared valence pattern).
\end{enumerate}

\section{Case Studies}
\label{sec:cases}

We illustrate the use of the FrameNet-based API to GF RGL by re-engineering two existing multilingual CNL grammars: one for translating standard tourist phrases \citep{RantaEtAl2012} and another for generating descriptions of paintings \citep{DannellsEtAl2012}, both developed in the MOLTO project.\footnote{\url{http://www.molto-project.eu/}} In both cases, we preserve the original functionality, and we do not make any changes in the application abstract syntax. Changes affect only the concrete syntaxes of English and Swedish.

\subsection{Phrasebook}
\label{ssec:phrasebook}

Although the Phrasebook grammar covers many idiomatic expressions that cannot be translated using the same frame or for which our approach would not be suitable as such, it includes around 20 complex clause-building functions that can be handled by the FN-based grammar. To illustrate the use of the FN-based grammar as a semantic API, we re-implement the following Phrasebook functions:

\begingroup
\fontsize{9pt}{11pt}\selectfont
\begin{verbatim}
ALive   : Person -> Country -> Action  -- e.g. `we live in Sweden'
AWant   : Person -> Object -> Action   -- e.g. `I want a pizza'
AWantGo : Person -> Place -> Action    -- e.g. `I want to go to a museum'
\end{verbatim}
\endgroup

\noindent by applying the frame functions $\mathsf{Desiring\_V2\_Act}$ and $\mathsf{Desiring\_VV}$ introduced in Section \ref{sec:fn-api}, and some additional functions:

\begingroup
\fontsize{9pt}{11pt}\selectfont
\begin{verbatim}
Motion_V_2    : Goal_Adv -> Source_Adv -> Theme_NP -> V -> Clause
Possession_V2 : Owner_NP -> Possession_NP -> V2 -> Clause
Residence_V   : Location_Adv -> Resident_NP -> V -> Clause
\end{verbatim}
\endgroup

By using RGL constructors, $\mathsf{ALive}$ is implemented for English, Swedish and other languages in the same way, except that different verbs are used:

\begingroup
\fontsize{9pt}{11pt}\selectfont
\begin{verbatim}
ALive p co = mkCl p.name (mkVP (mkVP (mkV "live")) (mkAdv in_Prep co))
ALive p co = mkCl p.name (mkVP (mkVP (mkV "bo")) (mkAdv in_Prep co))
\end{verbatim}
\endgroup

First, the language-specific verbs can be factored out by introducing a shared abstract verb in the domain lexicon (e.g. $\mathsf{live\_V}$ that links $\mathsf{live\_V\_Residence}$ and $\mathsf{bo\_V\_Residence}$). Second, the implementation of $\mathsf{ALive}$ can be done in a shared functor by using the FN-based API:

\begingroup
\fontsize{9pt}{11pt}\selectfont
\begin{verbatim}
ALive p co = let cl : Clause =
  Residence_V (Just Adv (mkAdv in_Prep co)) (Just NP p.name) live_V
    in mkCl cl.np cl.vp
\end{verbatim}
\endgroup

For $\mathsf{AWant}$, neither the original RGL-based nor the current FN-based implementation can be done in the functor because, in Swedish, the verb \emph{vilja} `to want' evoking $\mathsf{Desiring\_V2\_Act}$ requires the auxiliary verb \emph{ha} `to have'. This can be seen as a nested auxiliary frame \texttt{Possession}:

\begingroup
\fontsize{9pt}{11pt}\selectfont
\begin{verbatim}
AWant p obj = mkCl p.name (mkV2 (mkV "want")) obj       -- Eng
Desiring_V2_Act (Just NP p.name) (Just NP obj) want_V2
\end{verbatim}
\endgroup

\begingroup
\fontsize{9pt}{11pt}\selectfont
\begin{verbatim}
AWant p obj = mkCl p.name want_VV (mkVP L.have_V2 obj)  -- Swe
Desiring_VV
  (Just VP (Possession_V2 (Nothing NP) (Just NP obj) have_V2).vp)
  (Just NP p.name) want_VV
\end{verbatim}
\endgroup

Assuming that the auxiliary verb can be optionally used also with other Swedish verbs when applying this frame function, the nested frame could be hidden in the Swedish implementation of $\mathsf{Desiring\_V2\_Act}$. This, however, is not the case with $\mathsf{AWantGo}$ which in both languages requires a main nested frame and, thus, can be put in the functor:

\begingroup
\fontsize{9pt}{11pt}\selectfont
\begin{verbatim}
AWantGo p place = mkCl p.name want_VV (mkVP (mkVP go_V) place.to)
\end{verbatim}
\endgroup

\begingroup
\fontsize{9pt}{11pt}\selectfont
\begin{verbatim}
Desiring_VV (Just VP
  (Motion_V_2 (Just Adv place.to) (Nothing Adv) (Nothing NP) go_V).vp)
  (Just NP p.name) want_VV
\end{verbatim}
\endgroup

At the first gleam, the new code might look more complex, however, it does not specify how the verb phrases are built, and the same uniform code template is used in all cases.

The re-implemented version of Phrasebook accepts and generates the same set of sentences as before.\footnote{\url{http://www.grammaticalframework.org/demos/phrasebook/}}

\subsection{Paintings}
\label{ssec:museum}

The painting grammar is a part of a large-scale controlled NLG grammar developed for the cultural heritage domain in order to verbalize data about museum objects stored in an RDF-based ontology \citep{DannellsEtAl2012}. A set of RDF triples (subject-predicate-object expressions) forms the input to the application. As an example, a simplified set of triples representing information about the artwork \emph{Bacchus} is given below:

\begingroup
\fontsize{9pt}{11pt}\selectfont
\begin{verbatim}
<Bacchus> <createdBy> <Leonardo_da_Vinci>
<Bacchus> <hasDimension> <Bacchus_ImageDimesion>
<Bacchus> <hasCreationDate> <Bacchus_CreationDate>
<Bacchus> <hasCurrentLocation> <Musee_du_Louvre>
<Bacchus_ImageDimesion> <lengthValue> 115
<Bacchus_ImageDimesion> <heightValue> 177
<Bacchus_CreationDate> <timePeriodValue> 1510
\end{verbatim}
\endgroup

This information is combined by the grammar to generate a coherent text. A simplified abstract function that combines the triples is

\begingroup
\fontsize{9pt}{11pt}\selectfont
\begin{verbatim}
DPainting : Painting -> Painter -> Year -> Size -> Museum -> Description
\end{verbatim}
\endgroup

Each argument of the function corresponds to a class in the ontology. In Figure~\ref{fig:compare-paintings}, we show how the arguments are linearized in the original concrete syntax for English and how this syntax has been adapted to generate descriptions via the FN-based grammar. To adapt the original grammar, we first identified the frames that match the target verbs in the linearization rules. Then we matched the core FEs of the identified frames with the verb arguments.

Since the FN-based grammar currently does not cover non-core FEs, the adjunct {\sc Year} is associated with no FE in \texttt{Create\_physical\_artwork}. Instead, it is attached to the corresponding clause in the final linearization rule ($\mathsf{mkText}$), illustrating how non-core FEs can be incorporated.

\begin{figure}[t]
\begingroup
\fontsize{9pt}{11pt}\selectfont
\begin{verbatim}
DPainting painting painter           DPainting painting painter
 year size museum =                   year size museum =
let                                  let
 s1 : Text = mkText (mkS              cl1 : Clause =
  pastTense (mkCl painting (mkVP       Create_physical_artwork_V2_Pass
   (mkVP (passiveVP paint_V2)           (Just NP painter.long)
    (mkAdv by8agent_Prep                (Just NP painting)
     painter.long)) year.s))) ;         paint_V2 ;

 s2 : Text = mkText                   cl2 : Clause = Dimension_V2
  (mkCl it_NP (mkVP (mkVP              (Just NP (mkNP emptyNP size.s))
   (mkVPSlash measure_V2)              (Just NP it_NP)
   (mkNP (mkN ""))) size.s) ;          measure_V2 ;

 s3 : Text = mkText                   cl3 : Clause = Placing_V2_Pass
  (mkCl (mkNP this_Det painting)       (Just Adv museum.s)
   (mkVP (passiveVP display_V2)        (Just NP (mkNP this_Det painting))
    museum.s))                         display_V2

in mkText s1 (mkText s2 s3) ;        in mkText (mkText (mkS pastTense
                                      (mkCl cl1.np (mkVP cl1.vp year.s)))
                                      (mkText (mkCl cl2.np cl2.vp)
                                       (mkText (mkCl cl1.np cl3.vp))) ;
\end{verbatim}
\endgroup
\caption{An excerpt from the concrete painting grammar: before and after applying the FrameNet-based API (left column and right column respectively).}
\label{fig:compare-paintings}
\end{figure}

The grammar exploits patterns of frames \texttt{Create\_physical\_artwork}, \texttt{Dimension} and \texttt{Placing}:

\begingroup
\fontsize{9pt}{11pt}\selectfont
\begin{verbatim}
Create_physical_artwork_V2_Pass :
  Creator_NP -> Representation_NP -> V2 -> Clause
Dimension_V2 : Measurement_NP -> Object_NP -> V2 -> Clause
Placing_V2_Pass : Goal_Adv -> Theme_NP -> V2 -> Clause
\end{verbatim}
\endgroup

Alternatively, we could easily change the frame \texttt{Placing} with \texttt{Being\_located} evoked by the one-place verb \emph{hang} in the active voice, which would preserve the meaning but alter the linearization.

The Swedish syntax was adapted in the same way.
Descriptions generated by the new versions of \texttt{DPainting} are virtually equivalent to the descriptions produced by the original grammar.\footnote{\url{http://museum.ontotext.com/}}
The only difference in comparison to the original grammar is that in Swedish we have imposed the use of the main verb \emph{mäta} `to measure' instead of the copula:

\begin{quote}
Eng: \emph{Bacchus was painted by Leonardo da Vinci in 1510. It measures 115 by 177 cm. This work is displayed at the Musée du Louvre.}\\
Swe: \emph{Bacchus målades av Leonardo da Vinci år 1510. Den mäter 115 gånger 177 cm. Det här verket är utställt på Louvren.}
\end{quote}

\section{Evaluation}
\label{sec:evaluation}

We have conducted a simple intrinsic and extrinsic evaluation of the
acquired FN-based grammar and lexicon.
For an initial intrinsic evaluation, we count the number of examples
in the source corpora that belong to the set of shared frames and that
are covered by the shared semantico-syntactic valence patterns. Corpus
examples are judged by the sentence patterns that represent them,
disregarding non-core FEs, concrete prepositions and the word order,
but including syntactic roles and the grammatical voice. This means
that the original sentences are, in general, covered by paraphrasing.

We have extracted 57,615 examples from BFN and 3,348 examples from
SweFN that belong to the shared set of 483 frames. For both BFN and
SweFN, the concise set of 869 patterns covers 77.5\% of those
examples. This indicates that the set of shared patterns includes the
most frequently used ones despite the modest amount of the annotated example sentences in SweFN.

Based on the FN-annotated sentences covered by the shared valence
patterns, and the GF RGL type system for verbs, we have extracted
3,432 lexical entries (subcategorized LUs) from BFN,
and 1,899 entries form SweFN. LUs between BFN and SweFN are not
directly aligned, therefore a specific lexicon is generated for each
language. However, a partial shared lexicon has been automatically
derived on top of the language-specific lexicons, currently providing
a mapping between 703 LUs in BFN and 900 LUs in SweFN. The shared
lexicon covers 25.1\% (11,223) of BFN sentences and 35.8\% (928) of
SweFN sentences, counting only those sentences which are
represented by the shared valence patterns.

For an initial extrinsic evaluation, we compare the original
application grammars with their FN-based counterparts in terms of code
complexity. Since we do not modify the abstract syntax of application
grammars, the amount of linearization rules remains the
same. Therefore we count the number of constructors used to linearize
the functions. In the painting application, the number of constructors is
considerably reduced from 21 to 13. In the case of Phrasebook, the
number is slightly reduced from 10 in English and 11 in Swedish to 8
in both languages.

Another aspect of the evaluation with regard to the original application grammars is the large number of accurate high-level frame constructors which are available to the CNL application developers.
Instead of having to search for typical and valid syntactic patterns in a corpus to match the semantic representations of the application and domain, and to implement them, developers can choose among the abstract but still corpus-based semantico-syntactic patterns.
The frame semantics is consistent and can be mapped to the semantic representations of various applications in various domains having different levels of expressiveness.

\section{Discussion}
\label{sec:future}

The presented approach is based on
several assumptions that limit the scope of the shared grammar and
lexicon. The first difficulty is the low amount of annotated example sentences in SweFN. The differences
between the amounts of examples has become noticeable in the
set of extracted shared valence patterns. Without going into
further methodological details, we should note that the approach taken
in the development of SweFN was more lexicographically focused, putting
emphasis on enhancing frames with LUs rather than supplementing each
LU with example sentences. One way of adding more valence patterns for
verbs is from the morpho-syntactic descriptions provided in the
SIMPLE/PAROLE lexicons that are a part of the larger SweFN++ project.
These lexicons contain descriptive linguistic
analysis for around 3,000 Swedish verbs. Adding this information can
yield a larger, more representative set of shared valence
patterns additionally to the FN-annotated examples.

Furthermore, the extraction of verb valence patterns practically assumes
varied semantic descriptions, as well as large amounts of sentence
examples that are representative for the language in question. While
the BFN approach is likely to suggest frequent patterns and more general
linguistic descriptions, the SweFN approach is more likely to cover
the linguistic variation for each verb. The question of
how to balance between the two approaches has to be dealt with.

Another difficulty is selecting shared patterns in case of more than
two languages. Alternatives are: \begin{inparaenum}[(1)] \item an intersection of all
  languages, which means that the set of shared patterns inevitably gets
  smaller by adding each new language, but the intersection becomes
  more and more prototypical, provided that the corpora are of a
  reasonable size and coverage; \item a union of intersections of language
  pairs, which, on the one hand, would lead to functions temporary having no linearization in the one or the other language,
  but which, on the other hand, would be an efficient way to reveal non-compositional
  constructions and provide cross-lingual hints to the FN annotators and lexicographers.\end{inparaenum}

Non-compositional translation equivalents, when verb types differ or
when verbs do not have any counterpart in the other language, is yet another issue.
In SweFN, we find a range of verbs that lack an exact translation in English such
as: \emph{vabba} `to stay home because of a sick child', \emph{heta}
`be named', \emph{duka} `to make the table', \emph{diska} `to wash the
dishes', \emph{bädda} `to make the bed'. A related question here is to
what extent can these be represented in the grammar and how can we
represent them automatically. One possible solution is the reuse of
the GF RGL monolingual and multilingual dictionaries. Another solution is
finding complementary resources for constructing the FN-based lexicons, for
example by using WordNet for linking LUs.

For non-shared patterns and non-compositional translation equivalents, language specific extra modules can be
introduced. This will increase the coverage not only in monolingual applications but also in multilingual applications; however, it would require a manual, application-specific mapping between different frames.

The presented approach has some advantages with regard to GF
RGL. It can potentially provide feedback to the RGL monolingual and
multilingual dictionaries, yielding mutual benefits,
such as: \begin{inparaenum}[(1)] \item verification of verb
  types; \item verification of particle verbs; and \item suggestion of
  new entries. \end{inparaenum}

\section{Related Work}
\label{sec:related}

The main difference between this work and the previous approaches to CNL grammars is that we present an effort to exploit a robust and well established semantic model in the grammar development. Our approach can be compared with the work on multilingual verbalization of modular ontologies using GF and \emph{lemon}~\citep{DavisEtAl2012}, the Lexicon Model for Ontologies.
We use additional lexical information about syntactic arguments for building the concrete syntax.

The grounding of NLG using the frame semantics theory has been addressed in the work on text-to-scene generation~\citep{CoyneEtAl2011} and in the work on text generation for navigational tasks~\citep{RothAndFrank2010}.
In that research, the content of frames is utilized through alignment between the frame-semantic structure and the domain-semantic representation. Discourse is supported by applying aggregation and pronominalization techniques. In the cultural heritage use case, we also show how an application which utilizes the FN-based grammar can become more discourse-oriented; something that is necessary in actual NLG applications and that has been demonstrated in GF before~\citep{Dannells2010}.
In our current approach, the semantic representation of the domain and the linguistic structures of the grammar are based on FN-annotated data.

As suggested before~\citep{GruzitisAndBarzdins2010}, a FN-like approach can be used to deal with polysemy in CNL texts. Although we consider lexicalisation alternatives and restrictions for LUs and FEs, we do not address the problem of selectional restrictions and word sense disambiguation in general.

\section{Conclusion}
\label{sec:conclusion}

In this article, we presented a computational approach to multilingual
grammar and lexicon extraction and generation from FN-annotated
corpora. The methodology for constructing the grammars and the
lexicons was evaluated in a series of experiments. The results show that we are
able to generalize over a set of valence patterns to capture the
semantics and the syntax of two languages having a shared FN-based abstract syntax.
We discussed a number of potential improvements to achieve better results that would lead to a larger
coverage, however, the current coverage is already of practical use.
We have tested the feasibility of the generated grammar library as a semantic
API for developing CNL applications in GF. The major advantage is that
language-dependent clause-level specifications to a large extent are
hidden by the semantic API, making the application grammars more robust and
flexible.

\bibliographystyle{spbasic} 
\bibliography{framenet-gf} 

\end{document}